\documentclass[11pt]{article}

\usepackage[final]{acl}

\usepackage{times}
\usepackage{latexsym}
\usepackage{xcolor}
\usepackage[table]{xcolor}
\usepackage{tcolorbox}
\usepackage{placeins}
\usepackage[T1]{fontenc}

\usepackage[utf8]{inputenc}
\usepackage{booktabs}
\usepackage{multirow}
\usepackage{multicol}
\usepackage{microtype}
\usepackage{subcaption}
\usepackage{inconsolata}
\usepackage{amsmath}
\usepackage{graphicx}
\usepackage{flushend}
\usepackage{caption}
\title{No One Fits All: From Fixed Prompting to Learned Routing in Multilingual LLMs}

\author{
    Wei-Chi Wu\textsuperscript{$\alpha$} \ 
    Sheng-Lun Wei\textsuperscript{$\alpha$}\ 
    Hen-Hsen Huang\textsuperscript{$\beta$}\ 
    Hsin-Hsi Chen\textsuperscript{$\alpha$$\gamma$}
    \\
    \textsuperscript{$\alpha$}Department of Computer Science and Information Engineering, \\
    National Taiwan University, Taiwan
    \\
    \textsuperscript{$\beta$}Institute of Information Science, Academia Sinica, Taiwan
    \\
    \textsuperscript{$\gamma$}AI Research Center (AINTU), National Taiwan University, Taiwan
    \\
    \texttt{wcwu@csie.ntu.edu.tw},\quad 
    \texttt{weisl@nlg.csie.ntu.edu.tw},
    \\
    \texttt{hhhuang@iis.sinica.edu.tw,\quad hhchen@ntu.edu.tw}
}

\begin{document}
\maketitle
\begin{abstract}
Translation-based prompting is widely used in multilingual LLMs, yet its effectiveness varies across languages and tasks. We evaluate prompting strategies across ten languages of different resource levels and four benchmarks. Our analysis shows that no single strategy is universally optimal. Translation strongly benefits low-resource languages even when translation quality is imperfect, high-resource languages gain little, and prompt-based self-routing underperforms explicit translation. Motivated by these findings, we formulate prompting strategy selection as a learned decision problem and introduce lightweight classifiers that predict whether native or translation-based prompting is optimal for each instance. The classifiers achieve statistically significant improvements over fixed strategies across four benchmarks and generalize to unseen task formats not observed during training. Further analysis reveals that language resource level, rather than translation quality alone, determines when translation is beneficial.
\end{abstract}

\section{Introduction}
Translation-based prompting, which translates inputs into English prior to inference, is a widely used strategy for multilingual large language models (LLMs) and often improves performance by leveraging stronger English-centric capabilities \cite{ghosh-etal-2025-survey}. However, recent studies show that this advantage is not universal. Native-language prompting can outperform translation-based approaches on culturally grounded tasks \cite{tam2025languagemattersmultilingualinput,nyandwi-etal-2025-grounding} and for models with reduced English bias \cite{liu-etal-2025-translation}. These findings challenge the assumption that translation into English is always beneficial, raising a fundamental question: \textit{when should translation be applied, and when is native-language prompting preferable?}

Prior work has largely focused on improving individual prompting paradigms rather than understanding or selecting between them. Methods such as QAlign~\cite{zhu-etal-2024-question} and mCoT~\cite{lai-nissim-2024-mcot} enhance translation-based prompting, while Strategic CoT~\cite{wang2024strategic} improves native language reasoning. However, these approaches implicitly assume a fixed prompting strategy and do not treat prompting strategy selection as a decision problem conditioned on the language and task pair.

This gap motivates three research questions.
(\textbf{RQ1}) \textit{Does one prompting strategy fit all languages and tasks?}
Through a systematic comparison across diverse languages and tasks, we find that no single strategy consistently dominates. Translation-based prompting benefits low-resource languages but often provides limited or no gains for high-resource languages, while prompt-based self routing yields only marginal improvements and underperforms explicit translation.
(\textbf{RQ2}) \textit{Can prompting strategy selection be learned?}
We formulate strategy selection as a learned decision problem and introduce a lightweight classifier that predicts whether simple native language or translation-based prompting, the two most iconic prompting strategies, is more effective for a given language and task pair. Consequently, the lightweight classifier consistently outperforms isolated strategies across models and task formats. 
(\textbf{RQ3}) \textit{Why does translation primarily benefit low-resource languages?}
We show that translation effectiveness is driven more by language resource level than translation quality alone, with the learned selector favoring translation for low-resource languages even when translation quality is imperfect.
In summary, our contributions are threefold: \textbf{1)} We present a systematic empirical study demonstrating that no single prompting strategy fits all languages and tasks. \textbf{2)} We introduce a decision-oriented framework for learned prompting strategy selection. \textbf{3)} We provide an analysis uncovering the central role of language resource level in determining when translation-based prompting is beneficial.

\begin{table*}[t!]
    \centering
    \small
    \begin{tabular}{l|cccc|ccc|ccc|c}
    \toprule
        \textbf{Prompt Method} & \textbf{ZH} & \textbf{ES} & \textbf{DE} & \textbf{HI} & \textbf{BN} & \textbf{ID} & \textbf{KO} & \textbf{SI} & \textbf{SW} & \textbf{YO} & \textbf{Avg} \\
        \midrule
        \textsc{Native} & 88.2 & 89.4 & 88.2 & 84.6 & 83.1 & 88.0 & 36.7 & 75.3 & 75.5 & 45.2 & 75.4 \\ 
        \textsc{Translate} & 87.5 & 89.0 & 88.4 & \textbf{86.2} & 85.8 & \textbf{88.4} & \textbf{86.6} & \textbf{82.7} & \textbf{81.0} & 64.0 & 84.0 \\ 
        \textsc{Sel-Trans} & \textbf{88.8} & 89.3 & 88.6 & 85.7 & 85.4 & 88.2 & \textbf{86.6} & 80.8 & 79.6 & \textbf{64.4} & 83.7 \\ 
        \textsc{SCoT-Native} & 85.8 & 86.5 & 84.3 & 83.2 & 71.3 & 82.5 & 77.6 & 71.0 & 65.2 & 46.4 & 75.4 \\ 
        \textsc{SCoT-Trans} & 88.3 & \textbf{89.5} & \textbf{88.9} & 85.7 & \textbf{87.1} & 88.3 & 86.4 & 81.1 & 80.9 & 63.5 & \textbf{84.0} \\ 
        \textsc{Prompt-Routing} & 87.5 & 89.2 & 88.2 & 85.5 & 85.2 & 88.0 & 78.5 & 81.7 & 80.9 & 62.8 & 82.8 \\ 
    \bottomrule
    \end{tabular}
    \caption{Accuracy (\%) of six prompting strategies across ten languages on Global-MMLU using Llama3.3-70B. Languages are grouped by resource level into high (ZH, ES, DE, HI), mid (BN, ID, KO), and low (SI, SW, YO).}
    \label{tab:prompt_evaluation_short}
\end{table*}

\section{Related Work}

\paragraph{Translation-Based Prompting.}
English chain-of-thought reasoning often outperforms native approaches due to 
English dominance in pretraining 
\cite{li2024languagerankermetricquantifying,kowtal-etal-2024-chain}. 
Recent methods improve the issue through question alignment 
\cite{zhu-etal-2024-question}, multilingual CoT reasoning 
\cite{lai-nissim-2024-mcot}, and instruction tuning with small set
\cite{shaham-etal-2024-multilingual}. Translation effectiveness correlates positively with quality, as low-quality translation can harm performance
\cite{liu-etal-2025-translation}.

\paragraph{Limitations and Alternatives.}
Translation fails for culturally grounded tasks 
\cite{tam2025languagemattersmultilingualinput,nyandwi-etal-2025-grounding}, 
models with reduced English bias \cite{liu-etal-2025-translation}, and 
certain task structures 
\cite{huang-etal-2023-languages,intrator-etal-2024-breaking}. Alternatives 
include Strategic CoT \cite{wang2024strategic} and Selective Translation 
\cite{kowtal-etal-2024-chain,mondshine-etal-2025-beyond,paul2025aligninglargelanguagemodels}. 
We learn to select between strategies, revealing that language resource level 
and response features, not translation quality alone, determine optimality.

\section{Experimental Setup}
\label{sec:expsetup}

\paragraph{Datasets and Languages.}
We primarily evaluate on Global-MMLU \cite{singh-etal-2025-global}, grouping languages by resource level into high (Chinese/ZH, Spanish/ES, German/DE,  Hindi/HI), mid (Bengali/BN, Indonesian/ID, Korean/KO), and low (Sinhala/SI, Swahili/SW, Yoruba/YO). For strategy selection, we use a 10\% training split with balanced language coverage and evaluate on the remaining 90\%. Generalization is assessed on MMLU-ProX \cite{xuan-etal-2025-mmlu} and out-of-domain benchmarks with different task formats: XQuAD \cite{Artetxe_2020}, mCSQA \cite{sakai-etal-2024-mcsqa}, and XCOPA \cite{ponti-etal-2020-xcopa}.

\paragraph{Prompting Strategies.}
We compare zero-shot native and translation-based prompting strategies, including \textsc{Native}, \textsc{Translate}, \textsc{Sel-Trans} \cite{mondshine-etal-2025-beyond}, Strategic CoT in native and English \cite{wang2024strategic}, and \textsc{Prompt-Routing}. Prompt templates and details are provided in Appendix~\ref{sec:appendixA.1}.

\paragraph{Models.}
Experiments are conducted using DeepSeek-v3.1 \cite{deepseekai2024deepseekv3technicalreport}, with additional strategy selection experiments on Llama-3.3-70B-Instruct \cite{llama3modelcard}. All models are used in zero-shot inference.

\paragraph{Learned Strategy Selection.}
We formulate strategy selection as a binary decision between \textsc{Native} and \textsc{Translate}. Training labels are assigned when exactly one strategy answers correctly; ambiguous cases are discarded. We train lightweight classifiers (XGBoost \cite{Chen_xgboost_2016}, MLP \cite{haykin1994neural}) using features capturing differences between native and translated inputs and responses. Details are in Appendix~\ref{app:classifier_settings}.

\paragraph{Features Engineering.} For each instance, we run both \textsc{NATIVE} and \textsc{TRANSLATE} to obtain responses ${r_n}$ and ${r_t}$, then extract features capturing their differences across four categories: (1) metadata, (2) question-level, (3) response-level, and (4) alignment. The same language-agnostic pipeline is applied uniformly to all instances. Complete feature definitions appear in Appendix~\ref{app:features}.

\definecolor{indomainbg}{RGB}{232, 245, 233}
\definecolor{outdomainbg}{RGB}{255, 243, 224}
\definecolor{oraclebg}{gray}{0.85}

\begin{table*}[ht!]
    \centering
    \small
    \setlength{\tabcolsep}{3.2pt}
    \begin{tabular}{@{}ll cccc ccc ccc c@{}}
    \toprule
        & & \multicolumn{4}{c}{\textit{High-Resource}} & \multicolumn{3}{c}{\textit{Mid-Resource}} & \multicolumn{3}{c}{\textit{Low-Resource}} & \\
        \cmidrule(lr){3-6} \cmidrule(lr){7-9} \cmidrule(lr){10-12}
        \textbf{Dataset} & \textbf{Method} & \textbf{ZH} & \textbf{ES} & \textbf{DE} & \textbf{HI} & \textbf{BN} & \textbf{ID} & \textbf{KO} & \textbf{SI} & \textbf{SW} & \textbf{YO} & \textbf{Avg} \\
    \midrule
        \rowcolor{indomainbg}
        & \textsc{Native} & 86.4 & 86.4 & 84.6 & 83.1 & 79.9 & 85.3 & 36.1 & 72.0 & 71.7 & 39.0 & 72.5 \\ 
        \rowcolor{indomainbg}
        & \textsc{Translate} & 86.0 & 86.8 & 85.5 & 84.4 & 83.0 & 86.0 & 84.8 & 80.0 & 79.2 & 61.6 & 81.7 \\ 
        \rowcolor{indomainbg}
        \multirow{-3}{*}{\textsc{Global-MMLU}} & \textsc{Classifier} (Ours) & \textbf{86.8} & \textbf{87.3} & \textbf{86.0} & \textbf{84.7} & \textbf{83.3} & \textbf{86.4} & \textbf{84.9} & \textbf{80.0} & \textbf{79.8} & \textbf{63.7} & \textbf{82.3} \\ 
        \rowcolor{oraclebg}
        & \textsc{Oracle} & 90.5 & 90.6 & 91.2 & 89.1 & 88.4 & 90.8 & 89.5 & 86.5 & 85.3 & 72.8 & 88.3 \\ 
    \cmidrule(lr){1-13}
        \rowcolor{indomainbg}
        & \textsc{Native} & 80.5 & 80.8 & 79.7 & 77.9 & 75.4 & 80.1 & 35.7 & -- & 69.3 & 43.4 & 69.2 \\ 
        \rowcolor{indomainbg}
        & \textsc{Translate} & 80.3 & 81.0 & 80.3 & 79.0 & 78.5 & 80.5 & \textbf{79.5} & -- & 75.6 & 64.5 & 77.7 \\ 
        \rowcolor{indomainbg}
        \multirow{-3}{*}{\textsc{MMLU-ProX}} & \textsc{Classifier} (Ours) & \textbf{80.8} & \textbf{81.3} & \textbf{80.5} & \textbf{79.2} & \textbf{78.7} & \textbf{80.7} & \textbf{79.5} & -- & \textbf{76.0} & \textbf{64.6} & \textbf{77.9} \\ 
        \rowcolor{oraclebg}
        & \textsc{Oracle} & 85.4 & 85.5 & 85.0 & 84.2 & 83.3 & 85.5 & 82.8 & -- & 80.6 & 70.8 & 82.6 \\ 
    \midrule
        \rowcolor{outdomainbg}
        & \textsc{Native} & 86.6 & 87.2 & 89.2 & 83.6 & -- & -- & -- & -- & -- & -- & 86.7 \\ 
        \rowcolor{outdomainbg}
        & \textsc{Translate} & 87.2 & \textbf{88.2} & \textbf{90.6} & 82.5 & -- & -- & -- & -- & -- & -- & 87.1 \\ 
        \rowcolor{outdomainbg}
        \multirow{-3}{*}{\textsc{XQuAD}} & \textsc{Classifier} (Ours) & \textbf{88.6} & 87.9 & 89.2 & \textbf{84.7} & -- & -- & -- & -- & -- & -- & \textbf{87.6} \\ 
        \rowcolor{oraclebg}
        & \textsc{Oracle} & 91.0 & 92.0 & 94.1 & 89.8 & -- & -- & -- & -- & -- & -- & 91.7 \\ 
    \cmidrule(lr){1-13}
        \rowcolor{outdomainbg}
        & \textsc{Native} & 27.6 & -- & 38.2 & -- & -- & -- & -- & -- & -- & -- & 32.9 \\ 
        \rowcolor{outdomainbg}
        & \textsc{Translate} & 28.2 & -- & 38.5 & -- & -- & -- & -- & -- & -- & -- & 33.4 \\
        \rowcolor{outdomainbg}
        \multirow{-3}{*}{\textsc{mCSQA}} & \textsc{Classifier} (Ours)& \textbf{28.4} & -- & \textbf{39.1} & -- & -- & -- & -- & -- & -- & -- & \textbf{33.8} \\ 
        \rowcolor{oraclebg}
        & \textsc{Oracle} & 36.8 & -- & 45.7 & -- & -- & -- & -- & -- & -- & -- & 39.9 \\ 
    \cmidrule(lr){1-13}
        \rowcolor{outdomainbg}
        & \textsc{Native} & 97.0 & -- & -- & -- & -- & 95.8 & -- & -- & 87.4 & -- & 93.4 \\ 
        \rowcolor{outdomainbg}
        & \textsc{Translate} & \textbf{97.4} & -- & -- & -- & -- & \textbf{96.8} & -- & -- & 91.8 & -- & 95.3 \\ 
        \rowcolor{outdomainbg}
        \multirow{-3}{*}{\textsc{XCOPA}} & \textsc{Classifier} (Ours) & \textbf{97.4} & -- & -- & -- & -- & 96.6 & -- & -- & \textbf{93.0} & -- & \textbf{95.7} \\
        \rowcolor{oraclebg}
        & \textsc{Oracle} & 99.0 & -- & -- & -- & -- & 98.6 & -- & -- & 97.0 & -- & 98.2 \\
    \bottomrule
    \end{tabular}
    \caption{Results on DeepSeek-v3.1 across in-domain (\colorbox{indomainbg}{green}) and out-of-domain (\colorbox{outdomainbg}{orange}) benchmarks. Best results are \textbf{bolded}; \colorbox{oraclebg}{\textsc{Oracle}} marks the upper bound where at least one of \textsc{Native} or \textsc{Translate} succeeds. Empty cells indicate languages not covered by the respective benchmark dataset, as detailed in Table~\ref{tab:dataset_list}.}
    \label{tab:classifier_accuracy_deepseek}
\end{table*}

\section{RQ1: Does One Strategy Fit All?}
\label{sec:prelim}

Building on prior findings that question the universality of translation-based prompting, we examine whether any single prompting strategy outperforms others across languages and tasks, as implied by a ``one-strategy-fits-all'' assumption.
Table~\ref{tab:prompt_evaluation_short} reveals three key findings. First, \textbf{no single strategy dominates}: while \textsc{SCoT-Trans} achieves the highest average (83.97\%), \textsc{Sel-Trans} wins for 3 languages (ZH, KO, YO) and \textsc{Translate} for 5 others (HI, ID, KO, SI, SW). Second, \textbf{resource level predicts strategy effectiveness}: low-resource languages consistently favor translation (SI/SW/YO: +5.5 to +18.8\% over native), while high-resource languages show the opposite trend (<1\%). Korean presents an extreme case with a 49.9\% gap between strategies, suggesting severe underrepresentation in training. Third, \textbf{prompt-based strategy selection fails}: \textsc{Prompt-Routing} (82.8\%) underperforms simple \textsc{Translate} (84.0\%), demonstrating that effective strategy selection requires learning from patterns rather than model self-assessment.

\section{RQ2: Can We Learn to Select?}
\label{sec:method}

\subsection{Problem Formulation}

For each question $q$ in language $\ell$, we generate responses using both 
\textsc{Native} ($r_n$) and \textsc{Translate} ($r_t$) strategies. Our goal 
is to train a binary classifier $f(q, r_n, r_t) \rightarrow \{0, 1\}$ that 
predicts which strategy yields the correct answer, where 0 selects \textsc{Native} 
and 1 selects \textsc{Translate}.

\subsection{Experiment Results}

Table~\ref{tab:classifier_accuracy_deepseek} reports the performance of the XGBoost classifier on DeepSeek-v3.1 across all benchmarks. Complete results for both DeepSeek-v3.1 and Llama-3.3-70B are provided in Appendix~\ref{app:classifier_results}.

\paragraph{In-domain Performance.}
On Global-MMLU test set, \textsc{Classifier} achieves 82.3\% accuracy, 
outperforming \textsc{Translate} (+0.6\%) and substantially exceeding 
\textsc{Native} (+9.8\%). The classifier captures 
consistent performance gains across all languages, especially with improvements most pronounced for YO (+2.1\% over best baseline). On MMLU-ProX, gains persist (about +0.2\% over best baseline), demonstrating robustness to increased difficulty.

\paragraph{Out-of-domain Generalization.}
Despite training only on multiple-choice questions, the classifier generalizes 
to different task formats. On XQuAD (extractive QA), it achieves 87.6\% (+0.5\% 
over best baseline). On XCOPA (causal reasoning), performance reaches 95.7\% 
(+0.4\%). Even on mCSQA's challenging examples, the classifier shows modest 
gains (33.8\% vs 33.4\%).

\begin{table*}[h]
\centering
\footnotesize
\setlength{\tabcolsep}{3pt}
\begin{tabular}{lll}
\toprule
 & \textbf{XGBoost} & \textbf{MLP} \\
\midrule
\multirow{3}{*}{DeepSeek-v3.1}
  & 1. Word Overlap (34.38\%)
  & 1. Word Overlap (35.10\%) \\
  & 2. Metadata (24.52\%)
  & 2. Response Quality (12.62\%) \\
  & 3. POS (9.97\%)
  & 3. Question-Level (10.55\%) \\
\midrule
\multirow{3}{*}{Llama-3.3-70B}
  & 1. Word Overlap (56.91\%)
  & 1. Word Overlap (36.78\%) \\
  & 2. Question-Level (24.62\%)
  & 2. POS (10.95\%) \\
  & 3. Response Quality (17.11\%)
  & 3. Question-Level (10.41\%) \\
\bottomrule
\end{tabular}
\caption{Top 3 most important feature groups (with importance scores, \%) for XGBoost and MLP classifiers on DeepSeek-v3.1 and Llama-3.3-70B.}
\label{tab:feature_importance_table}
\end{table*}
\begin{figure}[ht!]
    \centering
    \includegraphics[width=1\linewidth]{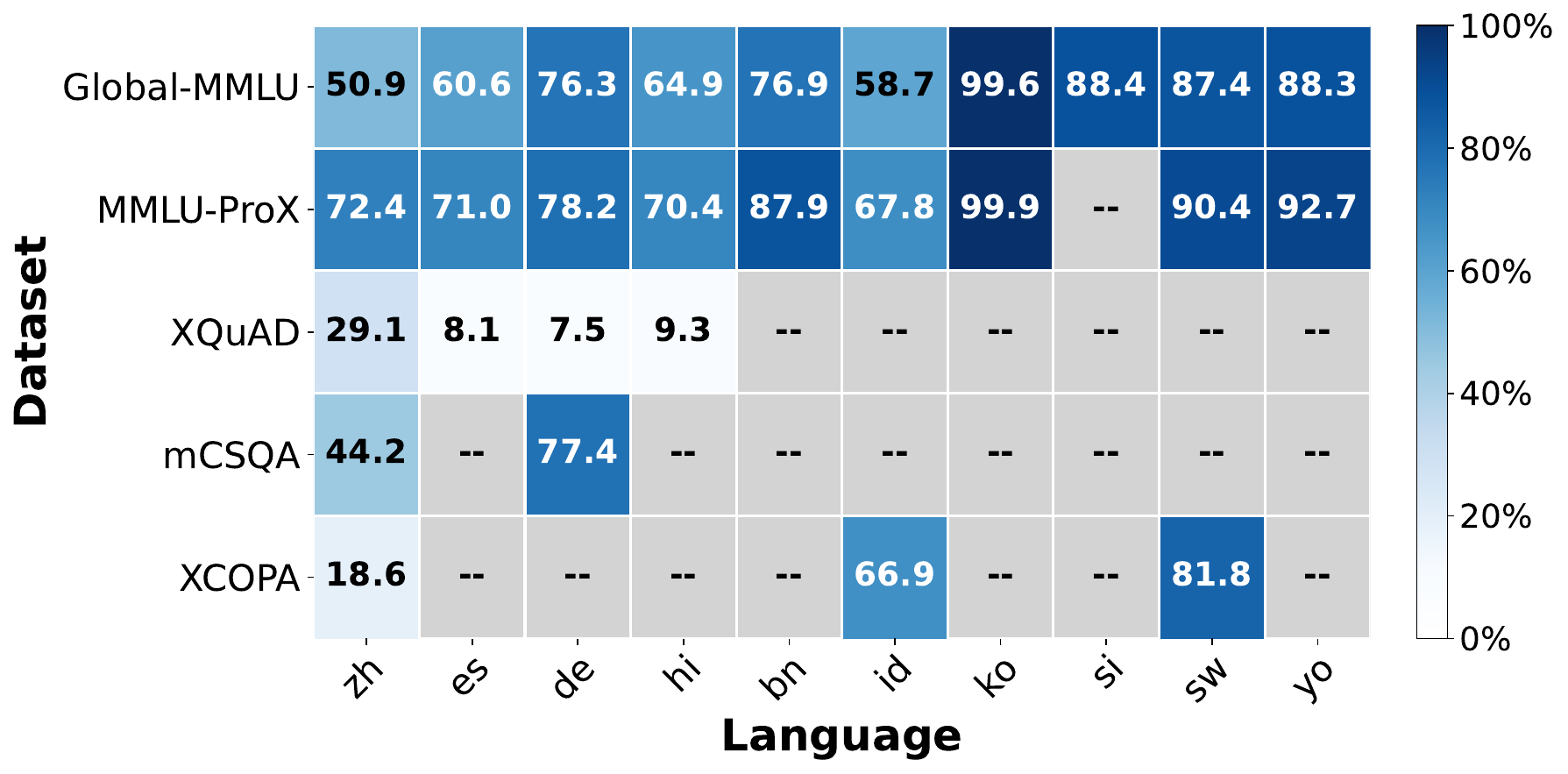}
    \caption{\textsc{Translate} selection rate (\%) of the XGBoost classifier on DeepSeek-v3.1.}
    \label{fig:transrate_heatmap}
\end{figure}

\paragraph{Statistical Significance.} 

We assess significance using the Wilcoxon signed-rank test~\cite{c4091bd3-d888-3152-8886-c284bf66a93a}. Across all language-dataset pairs on both models, XGBoost significantly outperforms both baselines ($p < 0.001$), while MLP achieves $p < 0.05$. 
This demonstrates the statistical robustness of learned strategy selection. Detailed calculation formula and results are presented in Appendix ~\ref{app:significance}.

\subsection{Feature Importance Analysis}
To further understand what drives the routing decisions, we analyze feature importance of the classifiers. Table~\ref{tab:feature_importance_table} shows that word overlap features consistently dominate across settings, suggesting that the classifiers primarily rely on semantic alignment differences between native and translated responses. 
These features are precisely what prompt-based self-routing cannot access. \textsc{PROMPT-ROUTING} relies on the model's self-assessment, which lacks the ability to quantify response-level differences and features. This explains why \textsc{PROMPT-ROUTING} (82.8\%) underperforms simple \textsc{TRANSLATE} (84.0\%), while the learned classifier, equipped with these features, consistently outperforms both fixed strategies.

\subsection{Strategy Selection Analysis}

Figure~\ref{fig:transrate_heatmap} shows the classifier's \textsc{Translate} selection rate strongly correlates with language resource level: relatively high-resource languages (ZH, ES, DE, HI, ID) exhibit balanced selection (40-70\%) varying by task, while relatively low-resource languages (KO, SI, YO) heavily favor \textsc{Translate}. This contradicts expectations from prior work, which indicates that translating low-resource languages leads to low translation quality~\cite{koehn-knowles-2017-six,nllbteam2022languageleftbehindscaling,shu2024transcendinglanguageboundariesharnessing}
and harm performance \cite{liu-etal-2025-translation}. We therefore investigate this relationship further in RQ3 (§\ref{sec:rq3}). Full translation rate heatmaps appear in Appendix ~\ref{app:transrate_all}.

\definecolor{lowqualitybg}{RGB}{255, 235, 238}   
\definecolor{midqualitybg}{RGB}{255, 248, 220}   
\definecolor{highqualitybg}{RGB}{232, 245, 233}  
\definecolor{correlbg}{gray}{0.85}               

\begin{table*}[h]
    \centering
    \small
    \setlength{\tabcolsep}{4pt}
    \begin{tabular}{@{}c cc cc cc cc cc@{}}
    \toprule
        & \multicolumn{2}{c}{\textit{Native}} & \multicolumn{2}{c}{\textit{Translate}} & \multicolumn{2}{c}{\textit{Classifier}} & \multicolumn{2}{c}{\textit{Gap (T-N)}} & \multicolumn{2}{c}{\textit{Trans Rate (\%)}} \\
        \cmidrule(lr){2-3} \cmidrule(lr){4-5} \cmidrule(lr){6-7} \cmidrule(lr){8-9} \cmidrule(lr){10-11}
        \textbf{Quality Percentile} & \textbf{DS} & \textbf{Llama} & \textbf{DS} & \textbf{Llama} & \textbf{DS} & \textbf{Llama} & \textbf{DS} & \textbf{Llama} & \textbf{DS} & \textbf{Llama} \\
    \midrule
        \rowcolor{lowqualitybg}
        10\% & 53.2 & 50.3 & 67.3 & 57.6 & 69.3 & 58.2 & 14.2 & 7.3 & 83.5 & 76.3 \\
        \rowcolor{lowqualitybg}
        20\% & 57.4 & 52.2 & 71.4 & 59.7 & 72.8 & 60.4 & 14.0 & 7.5 & 81.9 & 72.0 \\
        \rowcolor{lowqualitybg}
        30\% & 60.3 & 55.2 & 73.7 & 62.3 & 74.8 & 63.2 & 13.5 & 7.1 & 80.1 & 67.1 \\
        \rowcolor{midqualitybg}
        40\% & 62.1 & 57.7 & 75.0 & 64.4 & 75.9 & 65.3 & 12.9 & 6.7 & 78.6 & 63.2 \\
        \rowcolor{midqualitybg}
        50\% & 64.1 & 59.7 & 76.3 & 66.1 & 77.1 & 66.9 & 12.2 & 6.4 & 77.6 & 60.3 \\
        \rowcolor{midqualitybg}
        60\% & 65.8 & 61.3 & 77.4 & 67.4 & 78.1 & 68.3 & 11.6 & 6.0 & 76.6 & 58.0 \\
        \rowcolor{highqualitybg}
        70\% & 67.5 & 62.9 & 78.5 & 68.7 & 79.2 & 69.6 & 11.0 & 5.8 & 76.0 & 56.0 \\
        \rowcolor{highqualitybg}
        80\% & 69.2 & 64.3 & 79.5 & 69.8 & 80.1 & 70.7 & 10.3 & 5.6 & 75.5 & 53.8 \\
        \rowcolor{highqualitybg}
        90\% & 70.8 & 65.7 & 80.5 & 70.9 & 81.1 & 71.9 & 9.8 & 5.2 & 75.2 & 51.6 \\
        \rowcolor{highqualitybg}
        100\% & 72.5 & 67.2 & 81.7 & 72.1 & 82.3 & 73.1 & 9.3 & 4.9 & 75.2 & 49.2 \\
    \bottomrule
    \end{tabular}
    \caption{Translation quality analysis on Global-MMLU using chrF scores of the XGBoost classifier. 
    \colorbox{lowqualitybg}{Low quality} bins (bottom 30\%) show high \textsc{Translate} 
    selection rates despite lower accuracy. \colorbox{highqualitybg}{High quality} bins 
    (top 40\%) show improved accuracy but lower translation rate.
    }
    \label{tab:quality_main}
\end{table*}
\section{RQ3: Why Low-Resource Languages Favor Translation?}
\label{sec:rq3}

We conduct the analysis to explore the relationship among language resource level, translation quality, and learned strategy selection.

\paragraph{Setup.}
We evaluate translation quality using BLEURT \cite{sellam-etal-2020-bleurt}, 
chrF \cite{popovic-2015-chrf}, and METEOR \cite{banerjee-lavie-2005-meteor}, 
comparing model-generated translations against original English questions and options. 
We partition Global-MMLU and MMLU-ProX examples into quality deciles and 
measure: (1) accuracy for each method, (2) performance gap 
(\textsc{Translate} $-$ \textsc{Native}), and (3) classifier's 
\textsc{Translate} selection rate. Details appear in 
Appendix~\ref{app:quality_analysis}.

\paragraph{Results.}

Table~\ref{tab:quality_main} shows results using chrF on Global-MMLU 
for DeepSeek-v3.1 and Llama-3.3-70B. Complete results across quality metrics and datasets are provided in 
Appendix~\ref{app:quality_results_all}.
Three consistent patterns emerge. As translation quality improves: (1) all 
methods achieve higher accuracy, (2) the \textsc{Translate}$-$\textsc{Native} 
gap narrows, and (3) the classifier's \textsc{Translate} selection rate 
correspondingly decreases.
Critically, the classifier selects \textsc{Translate} most aggressively where 
translation quality is \textit{lowest}, not highest. This inverse correlation demonstrates the classifier learns to effectively
exploit translation where native performance is weakest, independent of translation quality itself.

\begin{figure}[ht!]
    \centering
    \includegraphics[width=1\linewidth]{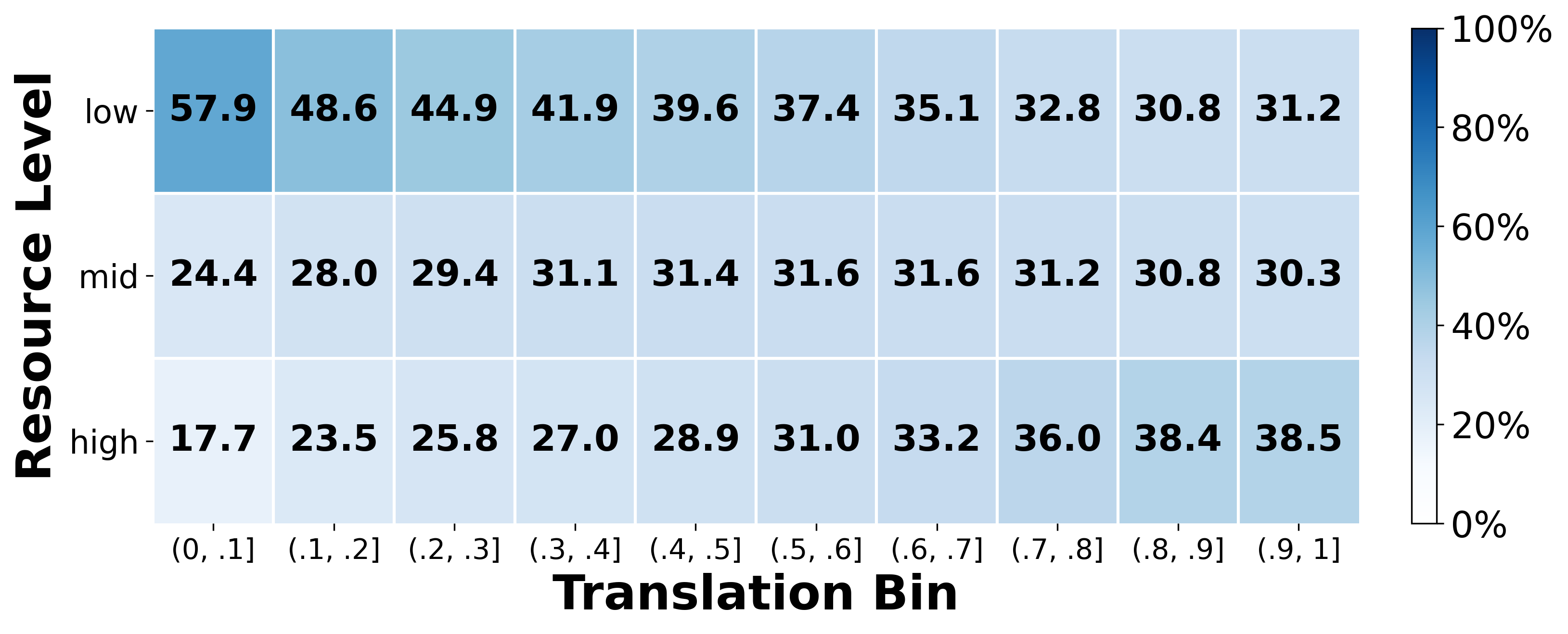}
    \caption{Distribution (\%) of responses across translation quality bins by language resource level on Global-MMLU with DeepSeek-v3.1. }
    \label{fig:resourcebin_heatmap}
\end{figure}

\paragraph{Discussion.}

This pattern reflects the confounding between language resource level, translation quality, and the learned strategy selection by classifiers.
As shown in Figure ~\ref{fig:resourcebin_heatmap}, the responses off low-resource languages concentrate in low-quality bins due to limited parallel corpora \cite{koehn-knowles-2017-six,nllbteam2022languageleftbehindscaling}, but 
benefit the most from translation (§\ref{sec:prelim}). 
The strategy performance gap narrows as 
high-resource languages dominate high-quality bins with 
little strategy differences. 
Our analysis reveals that \textbf{language resource level, not translation quality alone, determines optimal strategy}. Full language resource and quality bins heatmaps appear in Appendix ~\ref{app:quality_results_all}.

\section{Conclusion}
This work investigates prompting strategy selection for multilingual LLMs, showing that translation-based prompting is not universally beneficial and that no single strategy fits all language–task pairs, with low-resource languages favoring translation despite lower translation quality. To address this variability, we introduce lightweight classifiers that predict the optimal strategy for each instance, achieving statistically significant improvements over both native and translation baselines across four benchmarks and generalizing to unseen task formats. Through controlled analysis, we show that language resource level, rather than translation quality, is the primary factor determining when translation is beneficial. These findings reframe multilingual prompting from a fixed-strategy paradigm to a learned decision problem. Future work can build on this through stronger routing models, hybrid prompting strategies, retrieval-based selection methods, and ultimately integrating routing directly into model inference to eliminate dual-generation overhead.

\section*{Limitations}

While our classifier demonstrates effectiveness across multiple benchmarks, several limitations warrant consideration. First, our experiments focus on ten languages spanning different resource levels, and the generalizability of our findings to other unseen languages and additional model families, particularly non English-centric models, remains to be validated. Second, although our evaluation already covers a range of task types, it still does not fully represent the diversity of multilingual NLP applications due to the limitations of current existing multilingual datasets. Additionally, the lack of culturally sensitive multilingual datasets makes it difficult to assess whether cultural factors play a role in prompting strategies. Third, the classifier relies on features extracted from model responses, meaning it requires generating both native and translated outputs for each inference, which increases computational costs compared to selecting a single strategy. This overhead may limit practical deployment in resource-constrained settings. We expect that the strategy decision classifier could be further utilized within LLMs to automatically select the responding route without requiring dual inference, potentially through integration as an internal routing mechanism or by training the model to predict optimal strategies based on input features alone.

\section*{Acknowledgments}
This work was supported by National Science and Technology Council, Taiwan, under grant NSTC 114-2221-E-002 -070 -MY3, and by Ministry of Education (MOE), Taiwan, under grant NTU-114L900901.

\bibliography{anthology,custom}

\clearpage
\appendix

\section{Preliminary Details}
\subsection{LLM Endpoints}

We use the NVIDIA NIM APIs\footnote{\url{https://build.nvidia.com}} to generate responses for each prompting strategy, using \texttt{deepseek-ai/deepseek-v3\_1} with thinking mode enabled for DeepSeek-v3.1 and \texttt{meta/llama-3\_3-70b-instruct} for Llama-3.3-70B.

\subsection{Prompt Strategies}

\label{sec:appendixA.1}

We assess multiple strategies based on the language used for instructions and reasoning steps, using a zero-shot approach. The complete prompting templates are provided in Figure ~\ref{fig:prompt_example_native}, ~\ref{fig:prompt_example_translate}, ~\ref{fig:prompt_example_seltrans}, ~\ref{fig:prompt_example_scotnative}, ~\ref{fig:prompt_example_scottrans}, and ~\ref{fig:prompt_example_routing}. The prompt templates refer to \citet{liu-etal-2025-translation} (\textsc{Native}, \textsc{Translate}), \citet{mondshine-etal-2025-beyond} (\textsc{Sel-Trans}), and \citet{wang2024strategic} (\textsc{SCoT-Native}, \textsc{SCoT-Trans}).
 
\begin{figure*}
\begin{tcolorbox}[
title=\textit{Native Prompting}
]
\small
[Multiple-choice]\\
Answer the following multiple choice question. The last line of your response
should be exactly: 'Answer \$LETTER' where LETTER is one of ABCD.
Think step by step before answering.\\\\
Question: \{question\}\\\\
Options: \{options\}\\\\

[QA]\\
Answer the following question based on the given context. Provide a concise and accurate answer. The last line of your response should be exactly: 'Answer: [your answer]'.\\\\
Context: \{context\}\\\\
Question: \{question\}
\end{tcolorbox}
\caption{Native prompting template for LLM response generation. We use Google Translate to translate the instruction into other native languages when prompting.}
\label{fig:prompt_example_native}
\end{figure*}

\begin{figure*}
\begin{tcolorbox}[
title=\textit{Translate Prompting}
]
\small
[Multiple-choice] \\
First, translate the following question and options from \{language\} to English.
Then, answer the translated multiple choice question. 
The last line of your response should be exactly: 'Answer \$LETTER' where LETTER is one of ABCD.
Think step by step before answering.\\\\
Original Question (\{language\}): \{question\}\\\\
Original Options (\{language\}): \{options\}\\\\
Please provide your response in the following format: \\
Translated Question: [your English translation]\\
Translated Options: [your English translation]\\
Reasoning: [your step-by-step reasoning]\\
Answer [LETTER]\\\\

[QA]\\
First, translate the following context and question from \{language\} to English.
Then, answer the translated question based on the translated context.
The last line of your response should be exactly: 'Answer: [your answer]'.\\\\
Original Context (\{language\}): \{context\}\\\\
Original Question (\{language\}): \{question\}\\\\
Please provide your response in the following format:\\
Translated Context: [your English translation]\\
Translated Question: [your English translation]\\
Reasoning: [your step-by-step reasoning]\\
Answer: [your answer]
\end{tcolorbox}
\caption{Translate prompting template for LLM response generation. All languages use the same English instruction to translate and response.}
\label{fig:prompt_example_translate}
\end{figure*}

\begin{figure*}
\begin{tcolorbox}[
title=\textit{Selective-translate Prompting}
]
\small
[Multiple-choice]\\
Answer the following multiple choice question. The last line of your response
should be exactly: 'Answer \$LETTER' where LETTER is one of ABCD.
Think step by step before answering.\\\\
Question: \{question\}\\\\
Options: \{options\}
\end{tcolorbox}
\caption{Selective translate prompting template for LLM response generation. All languages use the same English instruction with native inputs.}
\label{fig:prompt_example_seltrans}
\end{figure*}

\begin{figure*}
\begin{tcolorbox}[
title=\textit{Native Strategic CoT Prompting}
]
\small
[Multiple-choice]\\
**Role:** You are a strategic reasoning expert skilled in systematic problem-solving.\\\\
**Workflow:**\\
1. First, analyze the problem and develop a strategic approach to solve it.\\
2. Then, apply your strategy step-by-step to reach the solution.\\\\
**Rule:** Ensure each step is logical, clear, and builds upon the previous one.\\\\
**Initialization:** Let's begin by understanding the problem and formulating a strategy.\\\\
**Task Input:**\\
Question: \{question\}\\\\
Options: \{options\}\\\\
Please follow the SCoT methodology: first outline your strategic approach, then apply it step-by-step. 
End your response with exactly: 'Answer \$LETTER' where LETTER is one of ABCD.
\end{tcolorbox}
\caption{Native strategic CoT prompting template for LLM response generation. We use Google Translate to translate the instruction into other native languages when prompting.}
\label{fig:prompt_example_scotnative}
\end{figure*}

\begin{figure*}
\begin{tcolorbox}[
title=\textit{Translate Strategic CoT Prompting}
]
\small
[Multiple-choice]\\
**Role:** You are a strategic reasoning expert skilled in systematic problem-solving.\\\\
**Workflow:**\\
1. First, analyze the problem and develop a strategic approach to solve it.\\
2. Then, apply your strategy step-by-step to reach the solution.\\\\
**Rule:** Ensure each step is logical, clear, and builds upon the previous one.\\\\
**Initialization:** Let's begin by understanding the problem and formulating a strategy.\\\\
**Task Input:**\\
Question: \{question\}\\\\
Options: \{options\}\\\\
Please follow the SCoT methodology: first outline your strategic approach, then apply it step-by-step. 
End your response with exactly: 'Answer \$LETTER' where LETTER is one of ABCD.
\end{tcolorbox}
\caption{Translate strategic CoT prompting template for LLM response generation. All languages use the same English instruction with native inputs.}
\label{fig:prompt_example_scottrans}
\end{figure*}

\begin{figure*}
\begin{tcolorbox}[
title=\textit{Routing Prompting}
]
\small
[Multiple-choice] \\
You are a multilingual AI assistant tasked with determining the best approach to answer a multiple-choice question.\\\\
Question Language: \{language\_name\}\\
Question: \{question\}\\
Options: \{options\}\\\\
Based on research in multilingual NLP, there are two approaches:\\
1. NATIVE: Answer directly in \{language\_name\}\\
2. TRANSLATE: Translate the question to English first, then answer\\\\
Please assess your proficiency and confidence:\\
- How confident are you in understanding and reasoning in \{language\_name\}? (Consider vocabulary, grammar, cultural context)\\
- Is this a complex question requiring nuanced reasoning, or is it straightforward?\\
- Would translating to English improve your accuracy?\\\\
Respond with EXACTLY ONE of the following on the last line:\\
ROUTE: NATIVE\\
or\\
ROUTE: TRANSLATE\\\\
Provide brief reasoning first (1-2 sentences), then your routing decision.\\\\

[QA]\\
You are a multilingual AI assistant tasked with determining the best approach to answer a question based on context.\\\\
Question Language: \{language\_name\}\\
Context: \{context\}\\
Question: \{question\}\\\\
Based on research in multilingual NLP, there are two approaches:\\
1. NATIVE: Answer directly in \{language\_name\} based on the context\\
2. TRANSLATE: Translate the context and question to English first, then answer\\\\
Please assess your proficiency and confidence:\\
- How confident are you in understanding and reasoning in \{language\_name\}? (Consider vocabulary, grammar, cultural context)\\
- Is this a complex question requiring nuanced reasoning, or is it straightforward?\\
- Would translating to English improve your accuracy?\\\\
Respond with EXACTLY ONE of the following on the last line:\\
ROUTE: NATIVE\\
or\\
ROUTE: TRANSLATE\\\\
Provide brief reasoning first (1-2 sentences), then your routing decision.
\end{tcolorbox}
\caption{Prompt routing template for LLM response generation. All languages use the same English instruction to decide the strategy. After the decision, they use the same prompt as native/translate prompting to get the final response and answer.}
\label{fig:prompt_example_routing}
\end{figure*}

\paragraph{Native method (\textsc{Native})} In \textsc{native}, we provide the question with both input and Chain-of-Thought instructions in the native language. 

\paragraph{Translate method (\textsc{Translate})} In \textsc{translate}, we provide the question with input in the native language, then instruct the model to translate the question to English and solve it with English Chain-of-Thought instructions.

\paragraph{Selective Translate method (\textsc{Sel-Trans})} \textsc{Selective Translation} \cite{mondshine-etal-2025-beyond} is a method that selectively translates only specific parts of the prompt. We provide the question with input in the native language, then instruct the model using English Chain-of-Thought instructions without first translating the question.

\paragraph{Native Strategic Chain-of-Thought method (\textsc{SCoT-Native})} \textsc{Native Strategic Chain-of-Thought} \cite{wang2024strategic} is a method that integrates strategic knowledge before generating intermediate reasoning steps. We provide both the input and Strategic Chain-of-Thought instructions in the native language.

\paragraph{Translate Strategic Chain-of-Thought method (\textsc{SCoT-Trans})} In \textsc{Translate Strategic Chain-of-Thought}, we provide the input in the native language, then instruct the model with English Strategic Chain-of-Thought instructions without first translating the complete question input.

\paragraph{Prompt Routing method (\textsc{Prompt-Routing})} In \textsc{Prompt-Routing}, we provide the input in the native language, then instruct the model to determine whether to translate the question into English and solve it with native or English Chain-of-Thought instructions.

\subsection{Complete Preliminary Analysis Results}

We conduct our preliminary analysis on the Global-MMLU subsets labeled as Culturally Sensitive (CS) and Culturally Agnostic (CA). The main results of the preliminary analysis separated by subsets 
are presented in Table \ref{tab:prompt_evaluation}.
\begin{table*}[!ht]
    \centering
    \small
    \begin{tabular}{ll|cccc|ccc|ccc|c}
    \toprule
        \textbf{Subset} & \textbf{Prompt Method} & \textbf{ZH} & \textbf{ES} & \textbf{DE} & \textbf{HI} & \textbf{BN} & \textbf{ID} & \textbf{KO} & \textbf{SI} & \textbf{SW} & \textbf{YO} & \textbf{Avg} \\
        \midrule
        \multirow{6}{*}{CS} & \textsc{Native} & 85.1 & 86.9 & 85.7 & 80.9 & 78.8 & 85.4 & 37.4 & 69.6 & 72.6 & 40.7 & 72.3 \\ 
         & \textsc{Translate} & 84.5 & \textbf{87.9} & 86.7 & \textbf{83.1} & 83.3 & \textbf{85.9} & \textbf{84.8} & \textbf{78.4} & 78.8 & \textbf{63.1} & \textbf{81.7} \\ 
         & \textsc{Sel-Trans} & \textbf{86.4} & 86.2 & 86.2 & 81.4 & 81.2 & 84.7 & 84.5 & 75.0 & 75.3 & 61.2 & 80.2 \\ 
         & \textsc{SCoT-Native} & 82.7 & 84.7 & 80.7 & 79.8 & 67.2 & 79.0 & 74.9 & 67.4 & 60.9 & 44.2 & 72.2 \\ 
         & \textsc{SCoT-Trans} & 86.1 & 87.0 & \textbf{87.1} & 82.1 & \textbf{87.1} & 84.3 & 84.0 & 77.0 & 76.9 & 62.2 & 81.4 \\ 
         & \textsc{Prompt-Routing} & 84.1 & 87.1 & 85.7 & 81.7 & 82.2 & 85.0 & 77.3 & 77.1 & \textbf{79.4} & 60.6 & 80.0 \\ 
        \midrule
        \multirow{6}{*}{CA} & \textsc{Native} & 89.4 & 90.4 & 89.2 & 86.1 & 84.8 & 89.0 & 36.5 & 77.5 & 76.6 & 47.0 & 76.7 \\ 
         & \textsc{Translate} & 88.6 & 89.5 & 89.0 & \textbf{87.4} & 86.7 & 89.3 & 87.3 & \textbf{84.3} & 81.8 & 64.4 & 84.8 \\ 
         & \textsc{Sel-Trans} & \textbf{89.7} & \textbf{90.5} & 89.5 & \textbf{87.4} & 87.0 & 89.5 & \textbf{87.4} & 83.0 & 81.3 & \textbf{65.6} & \textbf{85.1} \\ 
         & \textsc{SCoT-Native} & 87.0 & 87.2 & 85.7 & 84.5 & 72.9 & 83.8 & 78.7 & 72.4 & 66.9 & 47.2 & 76.6 \\ 
         & \textsc{SCoT-Trans} & 89.1 & 90.4 & \textbf{89.6} & 87.1 & \textbf{87.1} & \textbf{89.8} & 87.3 & 82.7 & \textbf{82.5} & 64.0 & 85.0 \\ 
         & \textsc{Prompt-Routing} & 88.8 & 90.0 & 89.1 & 86.9 & 86.3 & 89.1 & 78.9 & 83.4 & 81.4 & 63.6 & 83.8 \\ 
    \bottomrule
    \end{tabular}
    \caption{Accuracy (\%) of six prompting strategies on the Culturally Sensitive (CS) and Culturally Agnostic (CA) subsets of Global-MMLU using Llama3.3-70B, across ten languages grouped by resource level.}
    \label{tab:prompt_evaluation}
\end{table*}

\section{Classifiers Details}

\subsection{Classifier Settings and Details}
\label{app:classifier_settings}

We employ hyperparameter tuning to optimize the performance of our classifier models during the training session. We use Optuna~\cite{akiba2019optunanextgenerationhyperparameteroptimization} to perform automated hyperparameter optimization, 
optimizing overall accuracy (problem-level correctness) as the primary objective.
The final hyperparameter values selected for each model configuration are presented in Table~\ref{tab:hyperparameters}.

\paragraph{XGBoost} We tune the number of estimators (100--600), maximum tree depth (3--12), learning rate (0.01--0.3, log scale), subsample ratio (0.6--1.0), column subsample ratio (0.6--1.0), and minimum child weight (1.0--10.0).
\paragraph{MLP} We tune the hidden layer architecture (selected from predefined configurations), L2 regularization parameter $\alpha$ (1e-5--1e-2, log scale), and initial learning rate (1e-4--1e-2, log scale).

\begin{table*}[htbp]
    \centering
    \footnotesize
    \setlength{\tabcolsep}{3pt}
    \begin{tabular}{p{3.2cm}p{2.4cm}p{2.4cm}p{2.4cm}p{2.4cm}}
    \toprule
    \multirow{2}{*}{\textbf{Hyperparameter}} & \multicolumn{2}{c}{\textbf{MLP}} & \multicolumn{2}{c}{\textbf{XGBoost}} \\
    \cmidrule(lr){2-3} \cmidrule(lr){4-5}
    & \textbf{Deepseek-v3.1} & \textbf{Llama-3.3-70B} & \textbf{Deepseek-v3.1} & \textbf{Llama-3.3-70B} \\
    \midrule
    \multicolumn{5}{l}{\textit{MLP Hyperparameters}} \\
    \midrule
    Hidden Layer Sizes & $(100, 50)$ & $(100)$ & --- & --- \\
    $\alpha$ (L2 regularization) & $8.94 \times 10^{-5}$ & $4.19 \times 10^{-5}$ & --- & --- \\
    Learning Rate (initial) & $3.27 \times 10^{-3}$ & $5.44 \times 10^{-3}$ & --- & --- \\
    \midrule
    \multicolumn{5}{l}{\textit{XGBoost Hyperparameters}} \\
    \midrule
    Number of Estimators & --- & --- & 424 & 101 \\
    Max Depth & --- & --- & 10 & 3 \\
    Learning Rate & --- & --- & $2.87 \times 10^{-2}$ & $1.88 \times 10^{-2}$ \\
    Subsample & --- & --- & 0.951 & 0.700 \\
    Column Sample by Tree & --- & --- & 0.615 & 0.996 \\
    Min Child Weight & --- & --- & 9.51 & 4.71 \\
    \bottomrule
    \end{tabular}
    \caption{Final hyperparameter values for MLP and XGBoost classifiers.}\label{tab:hyperparameters}
\end{table*}

\subsection{Features}
\label{app:features}

Our approach relies on features capturing differences between $r_n$ and $r_t$ across linguistic quality, complexity, and alignment dimensions. Complete feature descriptions and examples appear in Table~\ref{tab:features}.

\paragraph{Metadata Features.}
Language identifier and subject category provide coarse-grained context about  resource availability and domain-specific requirements.

\paragraph{Question-Level Features.}
Punctuation mark count and numeric character count capture structural  properties that may interact differently with translation.

\paragraph{Response-Level Features.}
We compute linguistic quality metrics for both $r_n$ and $r_t$: named entity count using \texttt{spaCy} package, rare word ratio (words outside top-10k frequency), grammar fluency score, lexical diversity (type-token ratio), language confidence (probability assigned to detected language) using \texttt{langdetect} package, syntactic complexity (average dependency tree depth) using Stanza \cite{qi-etal-2020-stanza} models, and part-of-speech diversity using Stanza models.

\paragraph{Question-Response Alignment Features.}
Word overlap metrics and embedding similarity (cosine similarity of the combinations of question, answer, and response embeddings, using LaBSE \cite{feng-etal-2022-language}) measure how well each response addresses the question, potentially revealing translation-induced semantic drift.

\subsection{Training and Evaluation Datasets Details}

The complete statistics of the datasets is presented in Table ~\ref{tab:dataset_list}.

\subsection{Complete Classifier Results}
\label{app:classifier_results}

The complete accuracy results for the XGBoost and MLP classifier for DeepSeek-v3.1 and Llama-3.3-70B are presented in Table ~\ref{tab:classifier_accuracy_deepseek_all} and ~\ref{tab:classifier_accuracy_llama3.3_all}.

\section{Statistical Significance}
\label{app:significance}

The Wilcoxon signed-rank test evaluates whether the median difference between paired observations is zero. For each language-dataset pair $i$, we compute the difference $d_i = s_i^{\text{proposed}} - s_i^{\text{baseline}}$, where $s_i^{\text{proposed}}$ and $s_i^{\text{baseline}}$ are the accuracy scores of the proposed method and baseline, respectively. We then rank the absolute differences $|d_i|$ from smallest to largest, assigning rank $R_i$ to each pair. The test statistic $W$ is computed as:
\begin{align*}
    W = \min\left( \sum_{d_i > 0} R_i, \sum_{d_i < 0} R_i \right)
\end{align*}
where the first sum is over pairs where the proposed method outperforms the baseline, and the second sum is over pairs where the baseline performs better. Under the null hypothesis of no difference, $W$ follows a known distribution, from which we derive the $p$-value. The full result is presented in Table ~\ref{tab:wilcoxon_combined}.

\definecolor{oraclebg}{gray}{0.85}

\begin{table*}[htbp]
    \centering
    \footnotesize
    \begin{tabular}{p{4.5cm}p{5cm}p{4.5cm}}
    \toprule
    \textbf{Feature Name} & \textbf{Description} & \textbf{Example} \\
    \midrule
    \rowcolor{oraclebg}
    \multicolumn{3}{l}{\textit{Metadata Features}} \\
    \midrule
    \texttt{language} & Language code of the response (e.g., de, zh, es) & \texttt{"de"}, \texttt{"zh"} \\
    \texttt{dataset} & Name of the dataset (e.g., mmlu\_prox, xquad) & \texttt{"mmlu\_prox"} \\
    \texttt{subject} & Combined subject and category information & \texttt{"STEM:mathematics"} \\
    \midrule
    \rowcolor{oraclebg}
    \multicolumn{3}{l}{\textit{Question-Level Features}} \\
    \midrule
    \texttt{question\_punct\_density} & Density of punctuation marks in question text (punctuation count / text length) & Q: ``What is 2+2?'' (1 punct / 13 chars = 0.08) \\
    \texttt{question\_num\_density} & Density of numeric characters in question text (digit count / text length) & Q: ``What is 2+2?'' (2 digits / 13 chars = 0.15) \\
    \midrule
    \rowcolor{oraclebg}
    \multicolumn{3}{l}{\textit{Response-Level Features}} \\
    \midrule
    \texttt{rare\_word\_ratio} & Proportion of rare words based on corpus frequency (words below median frequency) & ``The method uses sophisticated techniques'' (rare words: sophisticated, techniques; 2 / 5 = 0.40) \\
    \texttt{named\_entity\_count} & Number of named entities (persons, organizations, locations) detected via \texttt{spaCy} & ``Einstein worked at Princeton in Germany'' (Einstein, Princeton, Germany = 3) \\
    \texttt{grammar\_fluency\_score} & Overall fluency score accounting for punctuation errors and formatting issues (0.0--1.0, higher is better) & ``The answer is correct.'' (1.0, no errors) vs ``The answer is correct..'' (0.82, malformed punctuation) \\
    \texttt{grammar\_malformed\_punct} & Count of malformed punctuation patterns (e.g., consecutive marks like "..") & ``Is this right??'' (1 instance of ``??'') \\
    \texttt{grammar\_missing\_final\_period} & Binary indicator of missing sentence-ending punctuation (1.0 = missing, 0.0 = present) & ``The answer is correct'' (1.0) vs ``The answer is correct.'' (0.0) \\
    \texttt{lexical\_diversity} & Type-token ratio measuring vocabulary diversity (unique words / total words) & ``The cat sat. The cat ran.'' (4 unique: the, cat, sat, ran / 6 total = 0.67) \\
    \texttt{language\_detection\_confidence} & Confidence score via \texttt{langdetect} (0.0--1.0, higher = more confident) & ``The quick brown fox jumps'' (detected as English with 0.95 confidence) \\
    \texttt{language\_mismatch} & Binary indicator of language mismatch via \texttt{langdetect} (1.0 = mismatch, 0.0 = match) & Expected: English, Detected: Spanish (1.0) \\
    \texttt{syntactic\_depth\_max} & Maximum depth of dependency parse tree (deeper = more complex) & ``The cat that the dog chased ran'' (deep nesting = depth 6) \\
    \texttt{syntactic\_complexity\_score} & Normalized syntactic complexity (depth / log2(word\_count + 1)) & ``The book that the student who the teacher praised read'' (depth 7, normalized) \\
    \texttt{pos\_noun\_verb\_ratio} & Ratio of nouns to verbs (higher = more nominal/informative style) & ``The analysis of the data shows results'' (4 nouns / 1 verb = 4.0) \\
    \texttt{pos\_diversity\_unique\_tags} & Number of unique part-of-speech tags in response & ``The cat sat on the mat'' (DET, NOUN, VERB, ADP = 4 unique tags) \\
    \texttt{pos\_diversity\_score} & POS diversity score (unique tags / total tags, 0.0--1.0) & ``The cat sat'' (3 unique / 3 total = 1.0) vs ``cat cat cat'' (1 unique / 3 total = 0.33) \\
    \midrule
    \rowcolor{oraclebg}
    \multicolumn{3}{l}{\textit{Alignment Features}} \\
    \midrule
    \texttt{word\_overlap\_*\_*} & Token-level overlap metrics (F1, precision, recall) measuring word overlap between pairs: answer--response, question--answer, and question--response & Response: ``The answer is Paris''; Reference: ``Paris''
    
    (overlap: \{paris\}, F1 = 0.33) \\
    \midrule
    \texttt{labse\_*\_*\_similarity} & Cosine similarity using LaBSE embeddings (cross-lingual semantic similarity) between pairs: answer--response, question--answer, and question--response & ``The capital is Paris'' (EN) vs ``La capital es París'' (ES) (similarity = 0.91) \\
    \bottomrule
    \end{tabular}
    \caption{Complete list of features used by the strategy selection classifier, grouped into four categories: metadata, question-level, response-level, and question–response alignment features.}
    \label{tab:features}
\end{table*}
    
\begin{table*}[!ht]
\small
\centering
\begin{tabular}{p{2.5cm}p{2.3cm}p{2.5cm}p{3cm}p{3.4cm}}
\hline
\textbf{Dataset} & \textbf{\# Examples} & \textbf{Used Languages} & \textbf{Task Type} & \textbf{Notes} \\
\hline
Global-MMLU & Around 14,000 per language & EN, ZH, ES, DE, HI, BN, ID, KO, SI, SW, YO & Multiple-choice with 4 options & We use 10\% of examples to train the classifier (§\ref{sec:expsetup}). \\
\hline
MMLU-ProX & Around 12,000 per language & EN, ZH, ES, DE, HI, BN, ID, KO, SW, YO & Harder multiple-choice format with more than 4 options &  \\
\hline
XQuAD & 1,190 per language & ZH, ES, DE, HI & Extractive QA & \\
\hline
mCSQA & 2,000-6,000 per language & ZH, DE & Commonsense QA & We only extract the examples with \texttt{hard} tag. \\
\hline
XCOPA & 500 per language & ZH, ID, SW & Causal reasoning &  \\
\hline
\end{tabular}
\caption{Statistics of the multilingual benchmark datasets used in our experiments, including the number of examples, covered languages, and task types.}
\label{tab:dataset_list}
\end{table*}
\definecolor{indomainbg}{RGB}{232, 245, 233}
\definecolor{outdomainbg}{RGB}{255, 243, 224}
\definecolor{oraclebg}{gray}{0.85}

\begin{table*}[!ht]
    \centering
    \small
    \setlength{\tabcolsep}{3.2pt}
    \begin{tabular}{@{}ll cccccccccc c@{}}
    \toprule
        & & \multicolumn{4}{c}{\textit{High-Resource}} & \multicolumn{3}{c}{\textit{Mid-Resource}} & \multicolumn{3}{c}{\textit{Low-Resource}} & \\
        \cmidrule(lr){3-6} \cmidrule(lr){7-9} \cmidrule(lr){10-12}
        \textbf{Dataset} & \textbf{Method} & \textbf{ZH} & \textbf{ES} & \textbf{DE} & \textbf{HI} & \textbf{BN} & \textbf{ID} & \textbf{KO} & \textbf{SI} & \textbf{SW} & \textbf{YO} & \textbf{Avg} \\
    \midrule
        \rowcolor{indomainbg}
        & \textsc{Native} & 86.4 & 86.4 & 84.6 & 83.1 & 79.9 & 85.3 & 36.1 & 72.0 & 71.7 & 39.0 & 72.5 \\ 
        \rowcolor{indomainbg}
        & \textsc{Translate} & 86.0 & 86.8 & 85.5 & 84.4 & 83.0 & 86.0 & 84.8 & 80.0 & 79.2 & 61.6 & 81.7 \\ 
        \rowcolor{indomainbg}
        & \textsc{XGBoost} (Ours) & \textbf{86.8} & \textbf{87.3} & \textbf{86.0} & \textbf{84.7} & \textbf{83.3} & \textbf{86.4} & \textbf{84.9} & \textbf{80.0} & \textbf{79.8} & \textbf{63.7} & \textbf{82.3} \\ 
        \rowcolor{indomainbg}
        \multirow{-4}{*}{\textsc{Global-MMLU}} & \textsc{MLP} (Ours) & 86.4 & 86.6 & 85.1 & 84.1 & 81.6 & 85.6 & 76.9 & 78.1 & 77.1 & 61.4 & 80.3 \\ 
        \rowcolor{oraclebg}
        & \textsc{Oracle} & 90.5 & 90.6 & 91.2 & 89.1 & 88.4 & 90.8 & 89.5 & 86.5 & 85.3 & 72.8 & 88.3 \\ 
    \midrule
        \rowcolor{indomainbg}
        & \textsc{Native} & 80.5 & 80.8 & 79.7 & 77.9 & 75.4 & 80.1 & 35.7 & -- & 69.3 & 43.4 & 69.2 \\ 
        \rowcolor{indomainbg}
        & \textsc{Translate} & 80.3 & 81.0 & 80.3 & 79.0 & 78.5 & 80.5 & \textbf{79.5} & -- & 75.6 & 64.5 & 77.7 \\ 
        \rowcolor{indomainbg}
        & \textsc{XGBoost} (Ours) & \textbf{80.8} & 81.3 & 80.5 & 79.2 & \textbf{78.7} & \textbf{80.7} & \textbf{79.5} & -- & \textbf{76.0} & 64.6 & \textbf{77.9} \\ 
        \rowcolor{indomainbg}
        \multirow{-4}{*}{\textsc{MMLU-ProX}} & \textsc{MLP} (Ours) & 80.8 & \textbf{81.3} & \textbf{80.7} & \textbf{79.3} & 78.0 & 79.3 & 78.2 & -- & 75.5 & \textbf{64.7} & 77.7 \\ 
        \rowcolor{oraclebg}
        & \textsc{Oracle} & 85.4 & 85.5 & 85.0 & 84.2 & 83.3 & 85.5 & 82.8 & -- & 80.6 & 70.8 & 82.6 \\ 
    \midrule
        \rowcolor{outdomainbg}
        & \textsc{Native} & 86.6 & 87.2 & 89.2 & 83.6 & -- & -- & -- & -- & -- & -- & 86.7 \\ 
        \rowcolor{outdomainbg}
        & \textsc{Translate} & 87.2 & \textbf{88.2} & \textbf{90.6} & 82.5 & -- & -- & -- & -- & -- & -- & 87.1 \\ 
        \rowcolor{outdomainbg}
        & \textsc{XGBoost} (Ours) & \textbf{88.6} & 87.9 & 89.2 & \textbf{84.7} & -- & -- & -- & -- & -- & -- & \textbf{87.6} \\ 
        \rowcolor{outdomainbg}
        \multirow{-4}{*}{\textsc{XQuAD}} & \textsc{MLP} (Ours) & 86.5 & 88.0 & 89.2 & 83.3 & -- & -- & -- & -- & -- & -- & 86.7 \\ 
        \rowcolor{oraclebg}
        & \textsc{Oracle} & 91.0 & 92.0 & 94.1 & 89.8 & -- & -- & -- & -- & -- & -- & 91.7 \\ 
    \midrule
        \rowcolor{outdomainbg}
        & \textsc{Native} & 27.6 & -- & 38.2 & -- & -- & -- & -- & -- & -- & -- & 32.9 \\ 
        \rowcolor{outdomainbg}
        & \textsc{Translate} & 28.2 & -- & 38.5 & -- & -- & -- & -- & -- & -- & -- & 33.4 \\
        \rowcolor{outdomainbg}
        & \textsc{XGBoost} (Ours) & \textbf{28.4} & -- & \textbf{39.1} & -- & -- & -- & -- & -- & -- & -- & \textbf{33.8} \\ 
        \rowcolor{outdomainbg}
        \multirow{-4}{*}{\textsc{mCSQA}} & \textsc{MLP} (Ours) & 28.4 & -- & 38.9 & -- & -- & -- & -- & -- & -- & -- & 33.7 \\ 
        \rowcolor{oraclebg}
        & \textsc{Oracle} & 36.8 & -- & 45.7 & -- & -- & -- & -- & -- & -- & -- & 39.9 \\ 
    \midrule
        \rowcolor{outdomainbg}
        & \textsc{Native} & 97.0 & -- & -- & -- & -- & 95.8 & -- & -- & 87.4 & -- & 93.4 \\ 
        \rowcolor{outdomainbg}
        & \textsc{Translate} & \textbf{97.4} & -- & -- & -- & -- & 96.8 & -- & -- & 91.8 & -- & 95.3 \\ 
        \rowcolor{outdomainbg}
        & \textsc{XGBoost} (Ours) & \textbf{97.4} & -- & -- & -- & -- & 96.6 & -- & -- & \textbf{93.0} & -- & \textbf{95.7} \\ 
        \rowcolor{outdomainbg}
        \multirow{-4}{*}{\textsc{XCOPA}} & \textsc{MLP} (Ours) & \textbf{97.4} & -- & -- & -- & -- & \textbf{97.0} & -- & -- & 89.6 & -- & 94.3 \\
        \rowcolor{oraclebg}
        & \textsc{Oracle} & 99.0 & -- & -- & -- & -- & 98.6 & -- & -- & 97.0 & -- & 98.2 \\ 
    \bottomrule
    \end{tabular}
    \caption{Complete results on DeepSeek-v3.1 across in-domain (\colorbox{indomainbg}{green}) and out-of-domain (\colorbox{outdomainbg}{orange}) benchmarks. Best results are \textbf{bolded}; \colorbox{oraclebg}{\textsc{Oracle}} marks the upper bound where at least one of \textsc{Native} or \textsc{Translate} succeeds. Empty cells indicate languages not covered by the respective benchmark dataset, as detailed in Table~\ref{tab:dataset_list}.}
    \label{tab:classifier_accuracy_deepseek_all}
\end{table*}

\begin{table*}[!ht]
    \centering
    \small
    \setlength{\tabcolsep}{3.2pt}
    \begin{tabular}{@{}ll cccccccccc c@{}}
    \toprule
        & & \multicolumn{4}{c}{\textit{High-Resource}} & \multicolumn{3}{c}{\textit{Mid-Resource}} & \multicolumn{3}{c}{\textit{Low-Resource}} & \\
        \cmidrule(lr){3-6} \cmidrule(lr){7-9} \cmidrule(lr){10-12}
        \textbf{Dataset} & \textbf{Method} & \textbf{ZH} & \textbf{ES} & \textbf{DE} & \textbf{HI} & \textbf{BN} & \textbf{ID} & \textbf{KO} & \textbf{SI} & \textbf{SW} & \textbf{YO} & \textbf{Avg} \\
    \midrule
        \rowcolor{indomainbg}
        & \textsc{Native} & 79.5 & 82.4 & \textbf{79.7} & 74.6 & 69.8 & 79.6 & 60.5 & 54.4 & 67.1 & 23.9 & 67.2 \\ 
        \rowcolor{indomainbg}
        & \textsc{Translate} & 78.5 & 80.6 & 76.9 & 76.2 & 72.9 & 78.6 & \textbf{76.5} & 67.0 & 69.8 & \textbf{43.0} & 72.0 \\ 
        \rowcolor{indomainbg}
        & \textsc{XGBoost} (Ours) & \textbf{80.0} & \textbf{82.9} & 79.5 & 76.7 & \textbf{73.4} & \textbf{80.9} & \textbf{76.5} & 68.1 & 69.7 & 43.0 & \textbf{73.1} \\
        \rowcolor{indomainbg}
        \multirow{-4}{*}{\textsc{Global-MMLU}} & \textsc{MLP} (Ours) & 79.8 & 82.5 & 79.7 & \textbf{76.8} & 73.2 & 80.2 & 69.8 & \textbf{69.1} & \textbf{70.3} & 40.7 & 72.2 \\
        \rowcolor{oraclebg}
        & \textsc{Oracle} & 85.3 & 87.2 & 85.7 & 83.4 & 81.3 & 85.3 & 82.5 & 76.0 & 79.3 & 55.1 & 80.1 \\
    \midrule
        \rowcolor{indomainbg}
        & \textsc{Native} & 63.8 & 66.0 & 65.4 & 57.8 & 53.4 & 65.5 & 40.2 & -- & 51.6 & 18.8 & 53.6 \\ 
        \rowcolor{indomainbg}
        & \textsc{Translate} & 57.8 & 59.9 & 58.7 & 54.5 & 52.9 & 59.6 & \textbf{56.0} & -- & 49.9 & \textbf{36.2} & 53.9 \\ 
        \rowcolor{indomainbg}
        & \textsc{XGBoost} (Ours) & 64.0 & \textbf{67.1} & 66.1 & 61.0 & 55.7 & \textbf{66.6} & 55.9 & -- & 54.2 & 36.0 & \textbf{58.5} \\ 
        \rowcolor{indomainbg}
        \multirow{-4}{*}{\textsc{MMLU-ProX}} & \textsc{MLP} (Ours) & \textbf{64.3} & 67.0 & \textbf{66.2} & \textbf{61.0} & \textbf{56.5} & \textbf{66.6} & 51.3 & -- & \textbf{56.1} & 35.8 & 58.3 \\
        \rowcolor{oraclebg}
        & \textsc{Oracle} & 71.3 & 72.3 & 71.4 & 67.3 & 64.4 & 72.0 & 63.9 & -- & 62.7 & 43.6 & 65.4 \\
    \midrule
        \rowcolor{outdomainbg}
        & \textsc{Native} & 86.1 & 87.4 & 88.4 & 84.5 & -- & -- & -- & -- & -- & -- & 86.6 \\ 
        \rowcolor{outdomainbg}
        & \textsc{Translate} & 83.9 & 86.6 & 88.3 & 80.1 & -- & -- & -- & -- & -- & -- & 84.7 \\ 
        \rowcolor{outdomainbg}
        & \textsc{XGBoost} (Ours) & 85.8 & \textbf{87.8} & \textbf{89.1} & \textbf{85.2} & -- & -- & -- & -- & -- & -- & \textbf{87.0} \\ 
        \rowcolor{outdomainbg}
        \multirow{-4}{*}{\textsc{XQuAD}} & \textsc{MLP} (Ours) & \textbf{86.1} & 87.3 & 88.8 & 84.5 & -- & -- & -- & -- & -- & -- & 86.7 \\
        \rowcolor{oraclebg}
        & \textsc{Oracle} & 90.3 & 91.9 & 92.4 & 88.4 & -- & -- & -- & -- & -- & -- & 90.8 \\ 
    \midrule
        \rowcolor{outdomainbg}
        & \textsc{Native} & 26.1 & -- & 33.8 & -- & -- & -- & -- & -- & -- & -- & 30.0 \\ 
        \rowcolor{outdomainbg}
        & \textsc{Translate} & 24.3 & -- & 33.5 & -- & -- & -- & -- & -- & -- & -- & 28.9 \\ 
        \rowcolor{outdomainbg}
        & \textsc{XGBoost} (Ours) & \textbf{26.5} & -- & 34.6 & -- & -- & -- & -- & -- & -- & -- & \textbf{30.6} \\ 
        \rowcolor{outdomainbg}
        \multirow{-4}{*}{\textsc{mCSQA}} & \textsc{MLP} (Ours) & 26.1 & -- & \textbf{35.2} & -- & -- & -- & -- & -- & -- & -- & 30.6 \\
        \rowcolor{oraclebg}
        & \textsc{Oracle} & 33.4 & -- & 41.8 & -- & -- & -- & -- & -- & -- & -- & 36.2 \\ 
    \midrule
        \rowcolor{outdomainbg}
        & \textsc{Native} & \textbf{97.8} & -- & -- & -- & -- & 96.6 & -- & -- & 83.0 & -- & 92.5 \\ 
        \rowcolor{outdomainbg}
        & \textsc{Translate} & 97.2 & -- & -- & -- & -- & \textbf{97.4} & -- & -- & \textbf{89.0} & -- & \textbf{94.5} \\ 
        \rowcolor{outdomainbg}
        & \textsc{XGBoost} (Ours) & 97.6 & -- & -- & -- & -- & 97.0 & -- & -- & 85.8 & -- & 93.5 \\
        \rowcolor{outdomainbg}
        \multirow{-4}{*}{\textsc{XCOPA}} & \textsc{MLP} (Ours) & \textbf{97.8} & -- & -- & -- & -- & 97.0 & -- & -- & 85.8 & -- & 93.5 \\
        \rowcolor{oraclebg}
        & \textsc{Oracle} & 99.0 & -- & -- & -- & -- & 98.0 & -- & -- & 94.4 & -- & 97.1 \\
    \bottomrule
    \end{tabular}
    \caption{Complete results on Llama3.3-70B across in-domain (\colorbox{indomainbg}{green}) and out-of-domain (\colorbox{outdomainbg}{orange}) benchmarks. Best results are \textbf{bolded}; \colorbox{oraclebg}{\textsc{Oracle}} marks the upper bound where at least one of \textsc{Native} or \textsc{Translate} succeeds. Empty cells indicate languages not covered by the respective benchmark dataset, as detailed in Table~\ref{tab:dataset_list}.}
    \label{tab:classifier_accuracy_llama3.3_all}
\end{table*}

\begin{table*}[!t]
    \centering
    \begin{tabular}{llc}
    \toprule
        \textbf{Model} & \textbf{Comparison} & \textbf{p-value} \\
    \midrule
        \multirow{4}{*}{DeepSeek-v3.1} & XGBoost vs Translate & $0.000873$ \\
        & XGBoost vs Native & $0.000006$ \\
        & MLP vs Translate & $0.037474$ \\
        & MLP vs Native & $0.000051$ \\
    \midrule
        \multirow{4}{*}{Llama3.3-70B} & XGBoost vs Translate & $0.000219$ \\
        & XGBoost vs Native & $0.000009$ \\
        & MLP vs Translate & $0.007202$ \\
        & MLP vs Native & $0.000035$ \\
    \bottomrule
    \end{tabular}
    \caption{Wilcoxon signed-rank test p-values combining all datasets.} \label{tab:wilcoxon_combined}
\end{table*}

\section{Translation Rate Results}

\subsection{Complete Translation Rate Heatmaps}
\label{app:transrate_all}

\begin{figure*}[t!]
    \centering
    \includegraphics[width=0.7\linewidth]{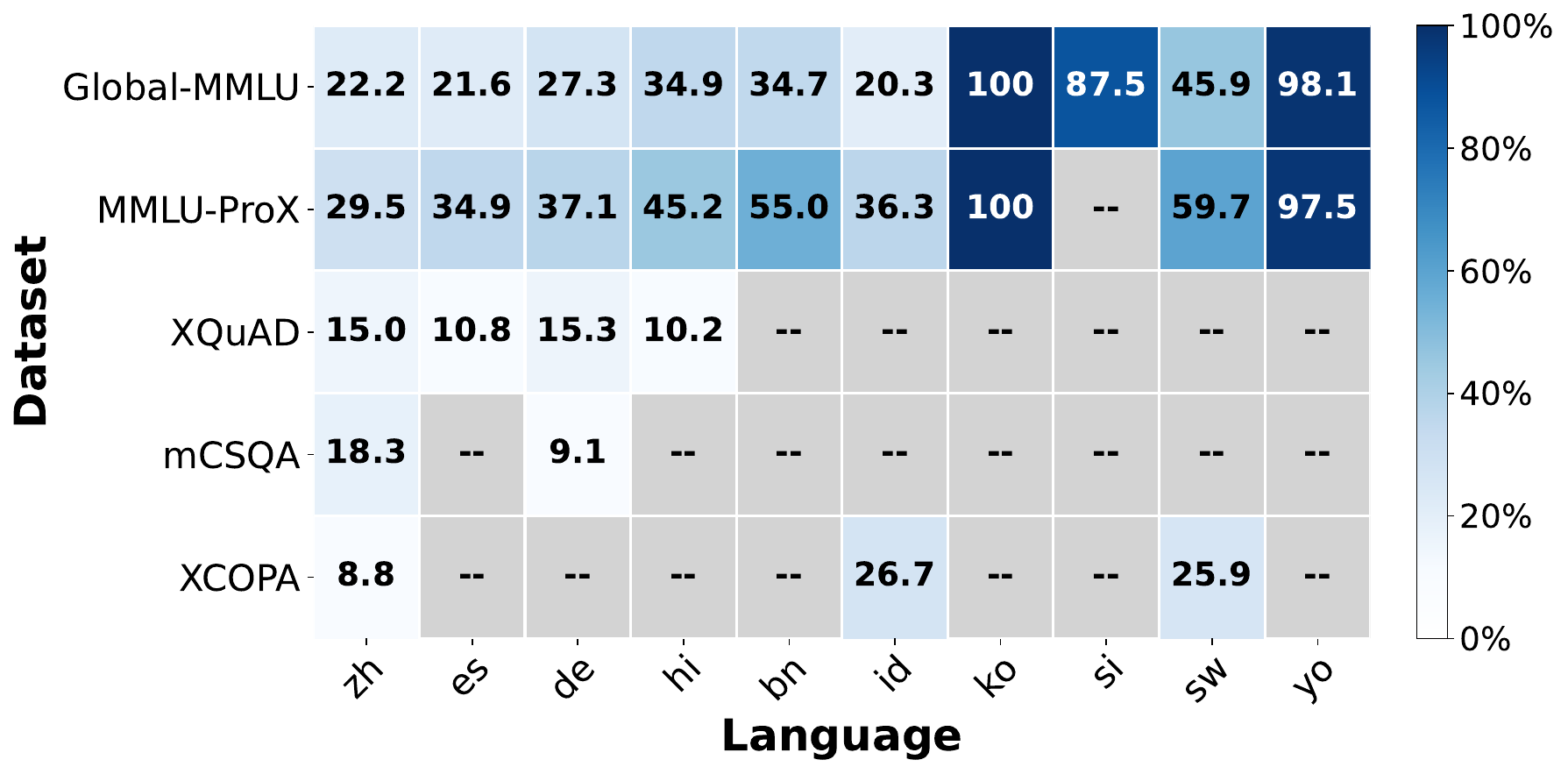}
    \caption{\textsc{Translate} selection rate (\%) of the XGBoost classifier on DeekSeek-v3.1.}
    \label{fig:transrate_heatmap_llama_xgboost}
\end{figure*}

\begin{figure*}[t!]
    \centering
    \includegraphics[width=0.7\linewidth]{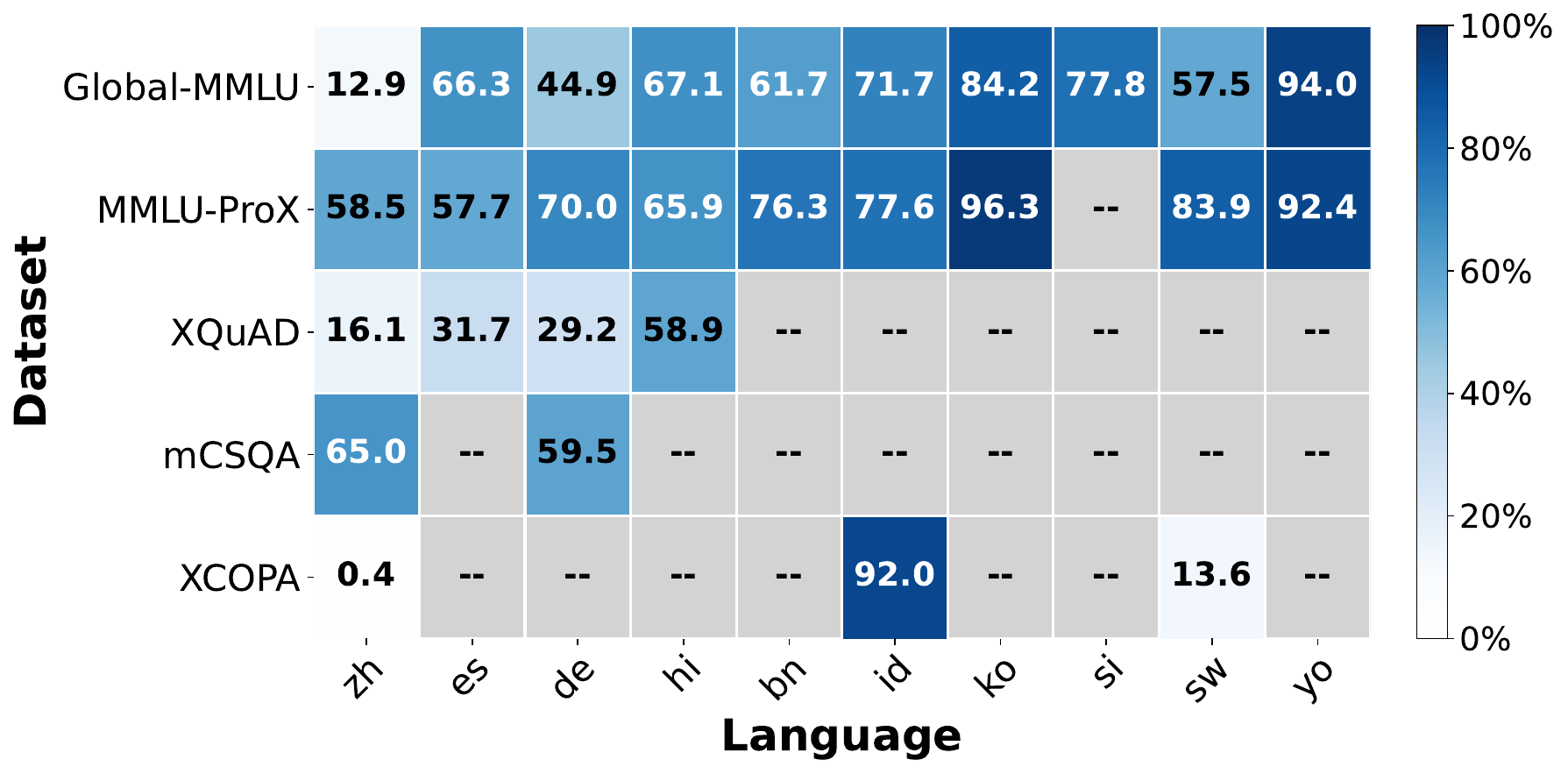}
    \caption{\textsc{Translate} selection rate (\%) of the MLP classifier on DeekSeek-v3.1.}
    \label{fig:transrate_heatmap_deepseek_mlp}
\end{figure*}

\begin{figure*}[t!]
    \centering
    \includegraphics[width=0.7\linewidth]{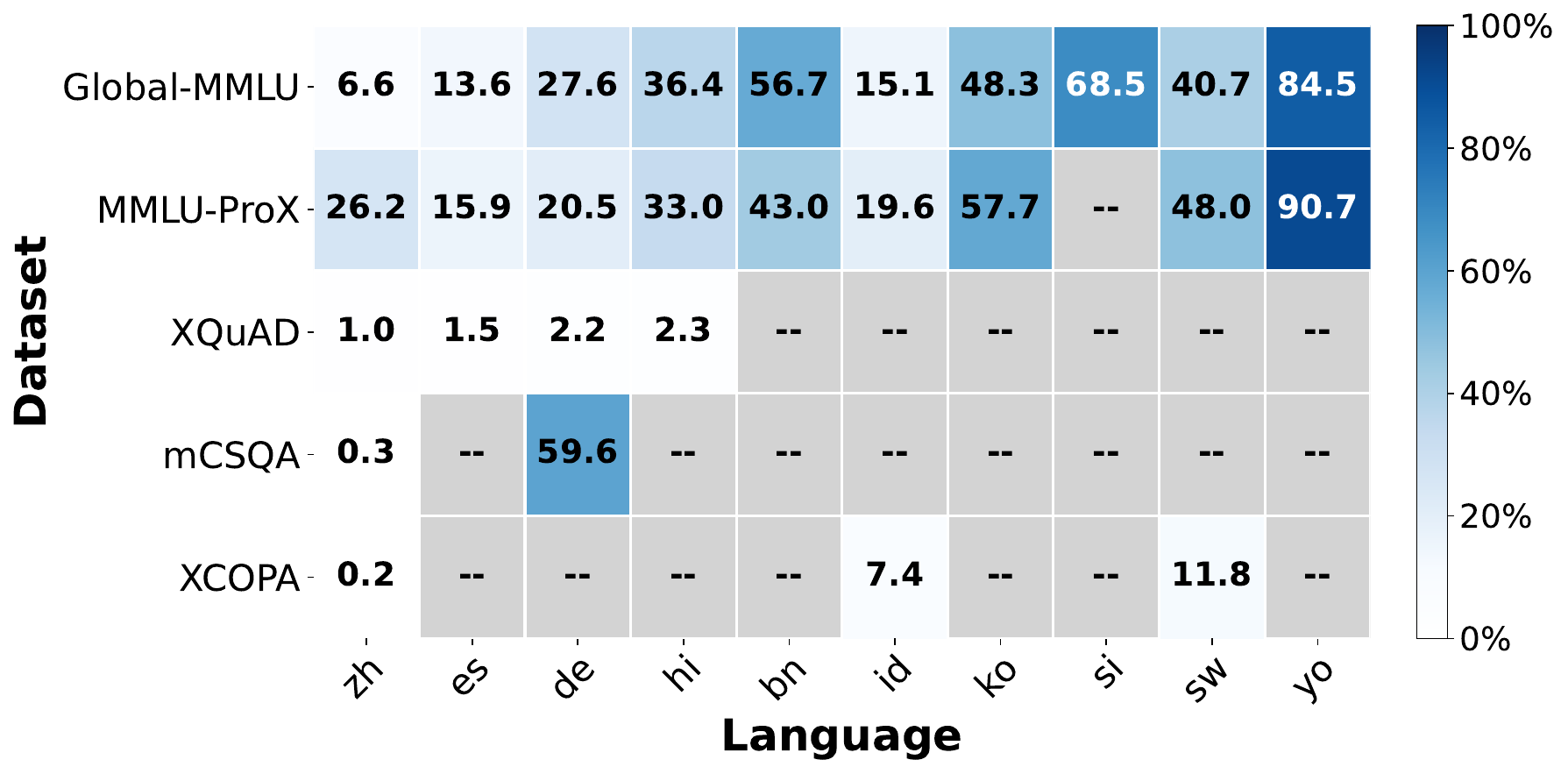}
    \caption{\textsc{Translate} selection rate (\%) of the MLP classifier on Llama-3.3-70B.}
    \label{fig:transrate_heatmap_llama_mlp}
\end{figure*}

The DeepSeek-v3.1 XGBoost classifier's translation rate heatmap is presented in Figure ~\ref{fig:transrate_heatmap} in the main body; the remaining classifiers 
translation rate heatmaps with different models and classifier's types are presented in Figure ~\ref{fig:transrate_heatmap_llama_xgboost}, ~\ref{fig:transrate_heatmap_deepseek_mlp}, and ~\ref{fig:transrate_heatmap_llama_mlp}.

\subsection{Translation Rate Analysis Experiment}
\label{app:quality_analysis}

We calculate all translation quality scores by comparing the translated question and options parsed from LLM responses to gold-standard English reference texts. Higher scores indicate translations that more faithfully preserve the meaning and structure of the original English content.

\paragraph{Results.}
\label{app:quality_results_all}

\begin{figure*}[!ht]
    \centering
    \includegraphics[width=1\linewidth]{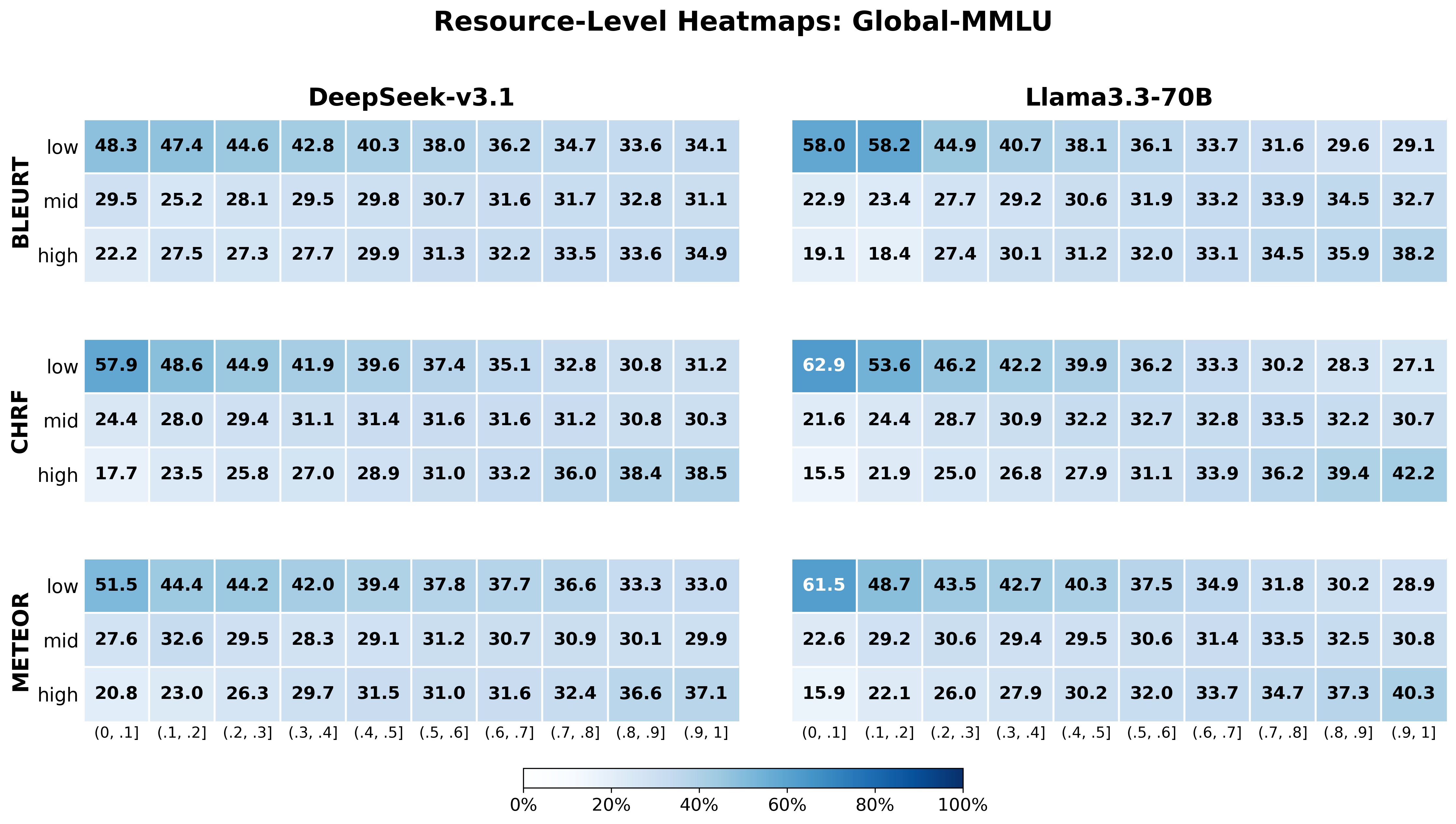}
    \caption{Combined distribution (\%) of responses across translation quality bins on Global-MMLU.}
    \label{fig:resource_bin_heatmap_combined_globalmmlu}
\end{figure*}

\begin{figure*}[!ht]
    \centering
    \includegraphics[width=1\linewidth]{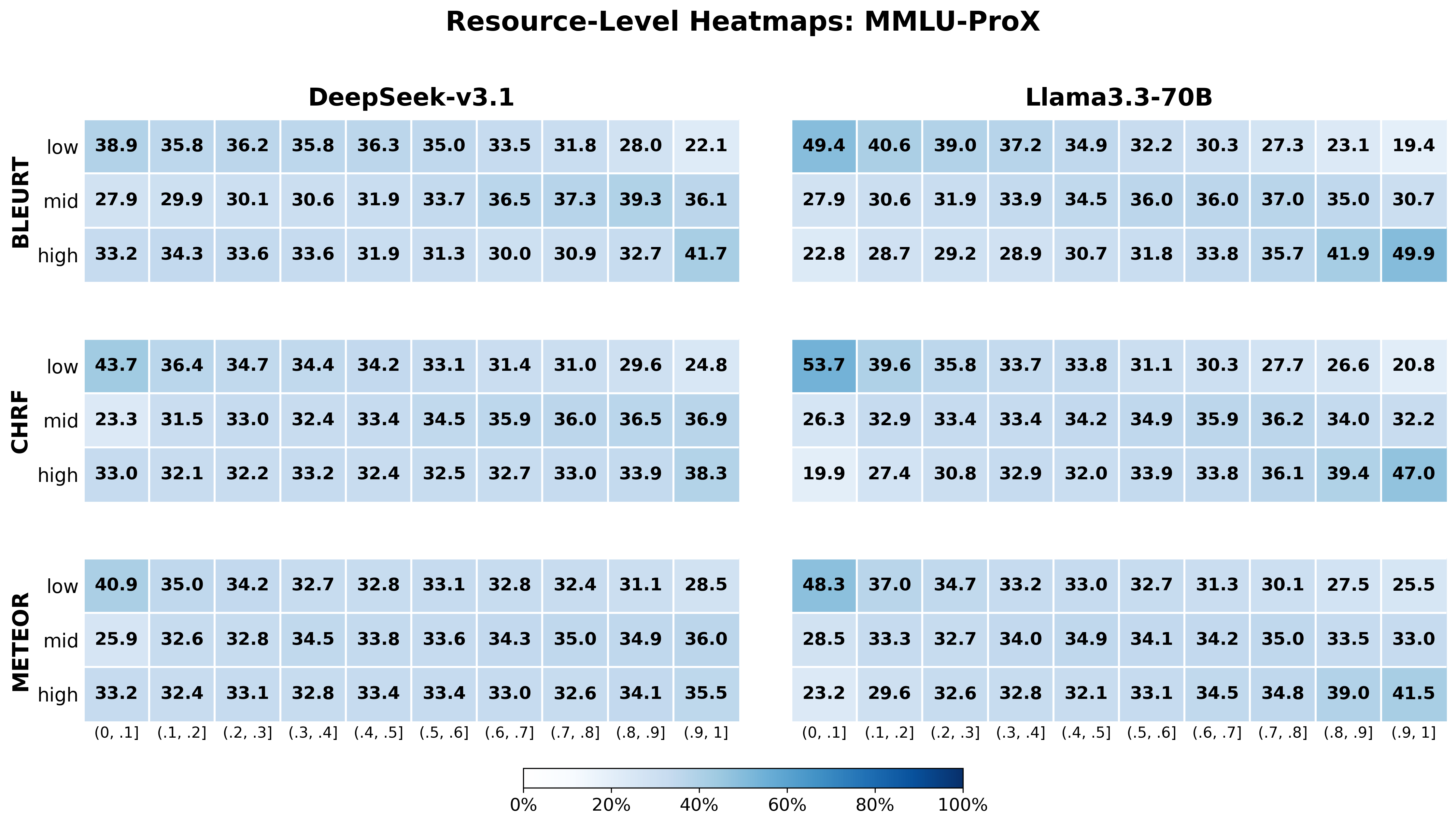}
    \caption{Combined distribution (\%) of responses across translation quality bins on MMLU-ProX.}
    \label{fig:resource_bin_heatmap_combined_mmluprox}
\end{figure*}

The Global-MMLU translation quality analysis results table is in Table ~\ref{tab:quality_main} in the main body; additional results of translation rate tables across three different quality scores, two datasets, and two models appear in Table ~\ref{tab:quality_globammlu_bleurt}, 
~\ref{tab:quality_globammlu_meteor}, 
~\ref{tab:quality_mmlu_prox_bleurt}, 
~\ref{tab:quality_mmlu_prox_chrf}, and 
~\ref{tab:quality_mmlu_prox_meteor}. Complete results of distribution of responses across translation quality bins are provided in Figure~\ref{fig:resource_bin_heatmap_combined_globalmmlu} and ~\ref{fig:resource_bin_heatmap_combined_mmluprox}.

\section{Information About Use Of AI Assistants}

We use AI assistants only for minimal tasks such as refining text and basic code snippets. All core research, experimental design, and critical examination of the results are performed and verified by humans to ensure the integrity of the process.

\clearpage

\begin{table*}[!ht]
    \centering
    \small
    \setlength{\tabcolsep}{4pt}
    \begin{tabular}{@{}c cc cc cc cc cc@{}}
    \toprule
        & \multicolumn{2}{c}{\textit{Native}} & \multicolumn{2}{c}{\textit{Translate}} & \multicolumn{2}{c}{\textit{Classifier}} & \multicolumn{2}{c}{\textit{Gap (T-N)}} & \multicolumn{2}{c}{\textit{Trans Rate (\%)}} \\
        \cmidrule(lr){2-3} \cmidrule(lr){4-5} \cmidrule(lr){6-7} \cmidrule(lr){8-9} \cmidrule(lr){10-11}
        \textbf{Quality Percentile} & \textbf{DS} & \textbf{Llama} & \textbf{DS} & \textbf{Llama} & \textbf{DS} & \textbf{Llama} & \textbf{DS} & \textbf{Llama} & \textbf{DS} & \textbf{Llama} \\
    \midrule
        \rowcolor{lowqualitybg}
        10\% & 55.9 & 53.8 & 68.4 & 60.0 & 69.7 & 60.6 & 12.5 & 6.2 & 77.5 & 71.9 \\
        \rowcolor{lowqualitybg}
        20\% & 58.2 & 52.2 & 70.7 & 59.5 & 72.0 & 60.2 & 12.5 & 7.3 & 77.0 & 71.0 \\
        \rowcolor{lowqualitybg}
        30\% & 60.5 & 54.7 & 72.9 & 61.9 & 73.9 & 62.6 & 12.4 & 7.2 & 76.2 & 67.4 \\
        \rowcolor{midqualitybg}
        40\% & 62.5 & 57.2 & 74.7 & 64.1 & 75.5 & 64.8 & 12.2 & 6.9 & 75.7 & 64.1 \\
        \rowcolor{midqualitybg}
        50\% & 64.3 & 59.4 & 76.1 & 66.0 & 76.9 & 66.7 & 11.8 & 6.6 & 75.2 & 61.1 \\
        \rowcolor{midqualitybg}
        60\% & 66.0 & 61.1 & 77.4 & 67.4 & 78.1 & 68.2 & 11.4 & 6.3 & 75.0 & 58.4 \\
        \rowcolor{highqualitybg}
        70\% & 67.7 & 62.7 & 78.5 & 68.7 & 79.1 & 69.5 & 10.8 & 6.0 & 74.7 & 56.1 \\
        \rowcolor{highqualitybg}
        80\% & 69.3 & 64.2 & 79.5 & 69.8 & 80.1 & 70.6 & 10.2 & 5.6 & 74.6 & 53.8 \\
        \rowcolor{highqualitybg}
        90\% & 70.9 & 65.7 & 80.6 & 70.9 & 81.2 & 71.9 & 9.7 & 5.2 & 74.8 & 51.5 \\
        \rowcolor{highqualitybg}
        100\% & 72.5 & 67.2 & 81.7 & 72.1 & 82.3 & 73.1 & 9.2 & 4.9 & 75.2 & 49.2 \\
    \bottomrule
    \end{tabular}
    \caption{Translation quality analysis on Global-MMLU using BLEURT scores. 
    \colorbox{lowqualitybg}{Low quality} bins (bottom 30\%) show high \textsc{Translate} 
    selection rates despite lower accuracy. \colorbox{highqualitybg}{High quality} bins 
    (top 40\%) show improved accuracy but lower translation rate.
    }
    \label{tab:quality_globammlu_bleurt}
\end{table*}

\begin{table*}[!ht]
    \centering
    \small
    \setlength{\tabcolsep}{4pt}
    \begin{tabular}{@{}c cc cc cc cc cc@{}}
    \toprule
        & \multicolumn{2}{c}{\textit{Native}} & \multicolumn{2}{c}{\textit{Translate}} & \multicolumn{2}{c}{\textit{Classifier}} & \multicolumn{2}{c}{\textit{Gap (T-N)}} & \multicolumn{2}{c}{\textit{Trans Rate (\%)}} \\
        \cmidrule(lr){2-3} \cmidrule(lr){4-5} \cmidrule(lr){6-7} \cmidrule(lr){8-9} \cmidrule(lr){10-11}
        \textbf{Quality Percentile} & \textbf{DS} & \textbf{Llama} & \textbf{DS} & \textbf{Llama} & \textbf{DS} & \textbf{Llama} & \textbf{DS} & \textbf{Llama} & \textbf{DS} & \textbf{Llama} \\
    \midrule
        \rowcolor{lowqualitybg}
        10\% & 52.5 & 50.5 & 68.2 & 57.9 & 70.0 & 58.6 & 15.7 & 7.4 & 80.0 & 75.3 \\
        \rowcolor{lowqualitybg}
        20\% & 56.6 & 52.1 & 71.3 & 59.8 & 72.5 & 60.4 & 14.7 & 7.7 & 78.8 & 71.9 \\
        \rowcolor{lowqualitybg}
        30\% & 59.5 & 55.1 & 73.2 & 62.2 & 74.2 & 63.0 & 13.7 & 7.1 & 77.5 & 67.3 \\
        \rowcolor{midqualitybg}
        40\% & 62.0 & 57.4 & 74.8 & 64.1 & 75.7 & 64.9 & 12.8 & 6.7 & 76.6 & 63.3 \\
        \rowcolor{midqualitybg}
        50\% & 64.1 & 59.5 & 76.2 & 65.8 & 77.0 & 66.7 & 12.1 & 6.3 & 75.9 & 60.3 \\
        \rowcolor{midqualitybg}
        60\% & 65.9 & 61.3 & 77.2 & 67.3 & 78.0 & 68.2 & 11.3 & 6.0 & 75.5 & 57.7 \\
        \rowcolor{highqualitybg}
        70\% & 67.6 & 62.8 & 78.3 & 68.6 & 79.0 & 69.5 & 10.7 & 5.8 & 75.1 & 55.4 \\
        \rowcolor{highqualitybg}
        80\% & 69.2 & 64.4 & 79.4 & 69.9 & 80.0 & 70.8 & 10.2 & 5.5 & 75.0 & 53.3 \\
        \rowcolor{highqualitybg}
        90\% & 70.8 & 65.8 & 80.6 & 71.0 & 81.1 & 72.0 & 9.8 & 5.2 & 74.9 & 51.4 \\
        \rowcolor{highqualitybg}
        100\% & 72.5 & 67.2 & 81.7 & 72.1 & 82.3 & 73.1 & 9.2 & 4.9 & 75.2 & 49.2 \\
    \bottomrule
    \end{tabular}
    \caption{Translation quality analysis on Global-MMLU using METEOR scores of the XGBoost classifier of the XGBoost classifier. 
    \colorbox{lowqualitybg}{Low quality} bins (bottom 30\%) show high \textsc{Translate} 
    selection rates despite lower accuracy. \colorbox{highqualitybg}{High quality} bins 
    (top 40\%) show improved accuracy but lower translation rate.}
    \label{tab:quality_globammlu_meteor}
\end{table*}

\definecolor{lowqualitybg}{RGB}{255, 235, 238}   
\definecolor{midqualitybg}{RGB}{255, 248, 220}   
\definecolor{highqualitybg}{RGB}{232, 245, 233}  

\begin{table*}[!ht]
    \centering
    \small
    \setlength{\tabcolsep}{4pt}
    \begin{tabular}{@{}c cc cc cc cc cc@{}}
    \toprule
        & \multicolumn{2}{c}{\textit{Native}} & \multicolumn{2}{c}{\textit{Translate}} & \multicolumn{2}{c}{\textit{Classifier}} & \multicolumn{2}{c}{\textit{Gap (T-N)}} & \multicolumn{2}{c}{\textit{Trans Rate (\%)}} \\
        \cmidrule(lr){2-3} \cmidrule(lr){4-5} \cmidrule(lr){6-7} \cmidrule(lr){8-9} \cmidrule(lr){10-11}
        \textbf{Quality Percentile} & \textbf{DS} & \textbf{Llama} & \textbf{DS} & \textbf{Llama} & \textbf{DS} & \textbf{Llama} & \textbf{DS} & \textbf{Llama} & \textbf{DS} & \textbf{Llama} \\
    \midrule
        \rowcolor{lowqualitybg}
        10\% & 62.0 & 42.7 & 71.9 & 45.8 & 72.4 & 48.4 & 9.9 & 3.1 & 85.8 & 63.7 \\
        \rowcolor{lowqualitybg}
        20\% & 64.1 & 46.7 & 73.3 & 49.0 & 73.7 & 52.4 & 9.2 & 2.3 & 85.2 & 59.3 \\
        \rowcolor{lowqualitybg}
        30\% & 65.3 & 48.1 & 74.5 & 50.0 & 74.8 & 53.9 & 9.2 & 1.9 & 85.0 & 58.2 \\
        \rowcolor{midqualitybg}
        40\% & 66.0 & 49.1 & 75.3 & 50.9 & 75.6 & 55.0 & 9.3 & 1.8 & 84.5 & 58.1 \\
        \rowcolor{midqualitybg}
        50\% & 66.1 & 49.6 & 75.5 & 51.2 & 75.7 & 55.6 & 9.4 & 1.6 & 84.0 & 58.1 \\
        \rowcolor{midqualitybg}
        60\% & 66.4 & 50.3 & 75.8 & 51.6 & 76.0 & 56.0 & 9.4 & 1.3 & 83.6 & 57.8 \\
        \rowcolor{highqualitybg}
        70\% & 66.6 & 51.0 & 76.0 & 52.1 & 76.3 & 56.5 & 9.4 & 1.1 & 83.1 & 57.5 \\
        \rowcolor{highqualitybg}
        80\% & 67.2 & 51.7 & 76.4 & 52.7 & 76.6 & 57.1 & 9.2 & 1.0 & 82.5 & 57.0 \\
        \rowcolor{highqualitybg}
        90\% & 68.0 & 52.6 & 76.9 & 53.2 & 77.2 & 57.7 & 8.9 & 0.6 & 81.8 & 56.1 \\
        \rowcolor{highqualitybg}
        100\% & 69.2 & 53.6 & 77.7 & 53.9 & 77.9 & 58.5 & 8.5 & 0.3 & 81.2 & 55.0 \\
    \bottomrule
    \end{tabular}
    \caption{Translation quality analysis on MMLU-ProX using BLEURT scores of the XGBoost classifier. 
    \colorbox{lowqualitybg}{Low quality} bins (bottom 30\%) show high \textsc{Translate} 
    selection rates despite lower accuracy. \colorbox{highqualitybg}{High quality} bins 
    (top 40\%) show improved accuracy but lower translation rate.}
    \label{tab:quality_mmlu_prox_bleurt}
\end{table*}

\begin{table*}[!ht]
    \centering
    \small
    \setlength{\tabcolsep}{4pt}
    \begin{tabular}{@{}c cc cc cc cc cc@{}}
    \toprule
        & \multicolumn{2}{c}{\textit{Native}} & \multicolumn{2}{c}{\textit{Translate}} & \multicolumn{2}{c}{\textit{Classifier}} & \multicolumn{2}{c}{\textit{Gap (T-N)}} & \multicolumn{2}{c}{\textit{Trans Rate (\%)}} \\
        \cmidrule(lr){2-3} \cmidrule(lr){4-5} \cmidrule(lr){6-7} \cmidrule(lr){8-9} \cmidrule(lr){10-11}
        \textbf{Quality Percentile} & \textbf{DS} & \textbf{Llama} & \textbf{DS} & \textbf{Llama} & \textbf{DS} & \textbf{Llama} & \textbf{DS} & \textbf{Llama} & \textbf{DS} & \textbf{Llama} \\
    \midrule
        \rowcolor{lowqualitybg}
        10\% & 60.0 & 43.5 & 71.8 & 46.5 & 72.3 & 48.5 & 11.8 & 3.0 & 81.6 & 70.7 \\
        \rowcolor{lowqualitybg}
        20\% & 61.6 & 48.2 & 73.0 & 50.8 & 73.5 & 53.1 & 11.4 & 2.6 & 80.4 & 65.5 \\
        \rowcolor{lowqualitybg}
        30\% & 62.7 & 50.2 & 73.6 & 52.1 & 74.1 & 54.8 & 10.9 & 1.9 & 79.8 & 62.2 \\
        \rowcolor{midqualitybg}
        40\% & 63.7 & 51.0 & 74.1 & 52.7 & 74.5 & 55.6 & 10.4 & 1.7 & 80.0 & 60.3 \\
        \rowcolor{midqualitybg}
        50\% & 64.6 & 51.3 & 74.8 & 52.8 & 75.2 & 56.0 & 10.2 & 1.5 & 80.3 & 59.2 \\
        \rowcolor{midqualitybg}
        60\% & 65.3 & 51.3 & 75.1 & 53.0 & 75.5 & 56.4 & 9.8 & 1.7 & 80.9 & 58.4 \\
        \rowcolor{highqualitybg}
        70\% & 65.9 & 51.6 & 75.6 & 53.0 & 75.9 & 56.7 & 9.7 & 1.4 & 81.2 & 57.7 \\
        \rowcolor{highqualitybg}
        80\% & 66.9 & 52.0 & 76.2 & 53.1 & 76.5 & 57.1 & 9.3 & 1.1 & 81.5 & 57.1 \\
        \rowcolor{highqualitybg}
        90\% & 68.0 & 52.6 & 76.9 & 53.2 & 77.2 & 57.6 & 8.9 & 0.6 & 81.6 & 56.4 \\
        \rowcolor{highqualitybg}
        100\% & 69.2 & 53.6 & 77.7 & 53.9 & 77.9 & 58.5 & 8.5 & 0.3 & 81.2 & 55.0 \\
    \bottomrule
    \end{tabular}
    \caption{Translation quality analysis on MMLU-ProX using chrF scores of the XGBoost classifier. 
    \colorbox{lowqualitybg}{Low quality} bins (bottom 30\%) show high \textsc{Translate} 
    selection rates on Llama-3.3-70B despite lower accuracy. \colorbox{highqualitybg}{High quality} bins 
    (top 40\%) show improved accuracy but lower translation rate.}
    \label{tab:quality_mmlu_prox_chrf}
\end{table*}

\begin{table*}[!ht]
    \centering
    \small
    \setlength{\tabcolsep}{4pt}
    \begin{tabular}{@{}c cc cc cc cc cc@{}}
    \toprule
        & \multicolumn{2}{c}{\textit{Native}} & \multicolumn{2}{c}{\textit{Translate}} & \multicolumn{2}{c}{\textit{Classifier}} & \multicolumn{2}{c}{\textit{Gap (T-N)}} & \multicolumn{2}{c}{\textit{Trans Rate (\%)}} \\
        \cmidrule(lr){2-3} \cmidrule(lr){4-5} \cmidrule(lr){6-7} \cmidrule(lr){8-9} \cmidrule(lr){10-11}
        \textbf{Quality Percentile} & \textbf{DS} & \textbf{Llama} & \textbf{DS} & \textbf{Llama} & \textbf{DS} & \textbf{Llama} & \textbf{DS} & \textbf{Llama} & \textbf{DS} & \textbf{Llama} \\
    \midrule
        \rowcolor{lowqualitybg}
        10\% & 61.5 & 47.3 & 73.9 & 50.1 & 74.4 & 52.0 & 12.4 & 2.8 & 79.8 & 68.7 \\
        \rowcolor{lowqualitybg}
        20\% & 62.2 & 50.4 & 73.7 & 52.4 & 74.2 & 54.5 & 11.5 & 2.0 & 78.2 & 63.7 \\
        \rowcolor{lowqualitybg}
        30\% & 62.1 & 50.8 & 72.9 & 52.3 & 73.4 & 54.9 & 10.8 & 1.5 & 78.3 & 61.6 \\
        \rowcolor{midqualitybg}
        40\% & 62.6 & 50.7 & 72.9 & 52.1 & 73.4 & 55.0 & 10.3 & 1.4 & 78.5 & 60.2 \\
        \rowcolor{midqualitybg}
        50\% & 63.8 & 50.8 & 73.8 & 52.2 & 74.2 & 55.3 & 10.0 & 1.4 & 79.2 & 59.3 \\
        \rowcolor{midqualitybg}
        60\% & 64.8 & 51.1 & 74.5 & 52.5 & 74.9 & 55.9 & 9.7 & 1.4 & 80.1 & 58.7 \\
        \rowcolor{highqualitybg}
        70\% & 66.0 & 51.6 & 75.4 & 52.8 & 75.7 & 56.5 & 9.4 & 1.2 & 80.9 & 57.8 \\
        \rowcolor{highqualitybg}
        80\% & 67.2 & 52.1 & 76.3 & 53.0 & 76.6 & 57.0 & 9.1 & 0.9 & 81.3 & 56.9 \\
        \rowcolor{highqualitybg}
        90\% & 68.3 & 52.8 & 77.1 & 53.3 & 77.3 & 57.7 & 8.8 & 0.5 & 81.5 & 56.1 \\
        \rowcolor{highqualitybg}
        100\% & 69.2 & 53.6 & 77.7 & 53.9 & 77.9 & 58.5 & 8.5 & 0.3 & 81.2 & 55.0 \\
    \bottomrule
    \end{tabular}
    \caption{Translation quality analysis on MMLU-ProX using METEOR scores of the XGBoost classifier. 
    \colorbox{lowqualitybg}{Low quality} bins (bottom 30\%) show different trend on two models on \textsc{Translate} 
    selection rates but both have lower accuracy. \colorbox{highqualitybg}{High quality} bins 
    (top 40\%) show improved accuracy, while DeepSeek-v3.1 model translation rate increases and Llama-3.3-70B model translation rate drops.}
    \label{tab:quality_mmlu_prox_meteor}
\end{table*}

\clearpage

\end{document}